\documentclass[manuscript]{acmart}
\usepackage{amssymb}

\usepackage{tabularx}
\newcolumntype{R}{>{\raggedleft\arraybackslash}X}%
\newcolumntype{C}{>{\centering\arraybackslash}X}%

\AtBeginDocument{%
  \providecommand\BibTeX{{%
    \normalfont B\kern-0.5em{\scshape i\kern-0.25em b}\kern-0.8em\TeX}}}

\setcopyright{acmcopyright}
\copyrightyear{2020}
\acmYear{2020}
\acmDOI{10.1145/1122445.1122456}

\usepackage{amsmath}

\begin{document}

%%
%% The "title" command has an optional parameter,
%% allowing the author to define a "short title" to be used in page headers.
\title{Evolving Context-Aware Recommender Systems With Users in Mind}

%%
%% The "author" command and its associated commands are used to define
%% the authors and their affiliations.
%% Of note is the shared affiliation of the first two authors, and the
%% "authornote" and "authornotemark" commands
%% used to denote shared contribution to the research.
\author{Amit Livne}
\authornote{Both authors contributed equally to this research.}
\email{livneam@post.bgu.ac.il}

\author{Eliad Shem Tov}
\authornotemark[1]
\email{eliads@post.bgu.ac.il}
\author{Adir Solomon}
\email{adirsolo@post.bgu.ac.il}
\author{Achiya Elyasaf}
\email{achiya@bgu.ac.il}
\author{Bracha Shapira}
\email{bshapira@bgu.ac.il}
\author{Lior Rokach}
\email{liorrk@bgu.ac.il}

\affiliation{%
  \institution{Ben-Gurion University of the Negev}
  \city{Beer-Sheva}
  \state{Israel}
}

%%
%% By default, the full list of authors will be used in the page
%% headers. Often, this list is too long, and will overlap
%% other information printed in the page headers. This command allows
%% the author to define a more concise list
%% of authors' names for this purpose.
\renewcommand{\shortauthors}{Livne and Shem Tov, et al.}

%%
%% The abstract is a short summary of the work to be presented in the
%% article.
\begin{abstract}
A context-aware recommender system (CARS) applies sensing and analysis of user context to provide personalized services. The contextual information can be driven from sensors in order to improve the accuracy of the recommendations. Yet, generating accurate recommendations is not enough to constitute a useful system from the users' perspective, since certain contextual information may cause different issues, such as draining the user's battery, privacy issues, and more. Additionally, adding high-dimensional contextual information may increase both the dimensionality and sparsity of the model.

Previous studies suggest reducing the amount of contextual information by selecting the most suitable contextual information using a domain knowledge. While in most studies the set of contexts is both small enough to handle and sufficient to prevent sparsity, such context sets do not necessarily represent an optimal set of features for the recommendation process. Another solution is compressing it into a denser latent space, thus disrupting the ability to explain the recommendation item to the user, and damaging users' trust.

In this paper we present an approach for selecting low-dimensional subsets of the contextual information and incorporating them explicitly within CARS. Specifically, we present a novel feature-selection algorithm, based on genetic algorithms (GA), that outperforms state-of-the-art dimensional-reduction CARS algorithms, improves the accuracy and the explainability of the recommendations, and allows for controlling user aspects, such as privacy and battery consumption. Furthermore, we exploit the top subsets that are generated along the evolutionary process, by learning multiple deep context-aware models and applying a stacking technique on them, thus improving the accuracy while remaining at the explicit space. 

We evaluated our approach on two high-dimensional context-aware datasets driven from smartphones. An empirical analysis of our results validates that our proposed approach outperforms state-of-the-art CARS models while improving transparency and explainability to the user. In addition to the empirical results, we provide many use cases and examples of how researchers and domain experts can tweak the feature-selection algorithm and use it for improving the user aspects and the explainability.
\end{abstract}

%%
%% The code below is generated by the tool at http://dl.acm.org/ccs.cfm.
%% Please copy and paste the code instead of the example below.
%%

%%
%% Keywords. The author(s) should pick words that accurately describe
%% the work being presented. Separate the keywords with commas.

%%
%% This command processes the author and affiliation and title
%% information and builds the first part of the formatted document.
\maketitle
\section{Introduction}
\label{sec:intro}
\emph{Recommendation Systems} (RS) are designed to assist users in locating suitable information or products and aid users in the decision making process. \emph{Context-Aware Recommender System} (CARS) is the sub-field of RS that studies the representation and integration of contextual information into RS. Contextual information can be used for inferring the specific situation under which recommendations are made and influences the recommendation goals~\cite{adomavicius2005incorporating,aggarwal2016recommender}. Contextual information can include auxiliary information such as time, weather, location, and more~\cite{unger2016towards,unger2018inferring}. The emergence and penetration of smart mobile devices have given rise to the development of context-aware systems that utilize sensors to collect data about users in order to improve and personalize services~\cite{perera2014context}. For example, by mining sensor data of a user's mobile, the system can determine the contextual situation of the user. Such information can aid personalizing the recommendations based on the user's dynamic and changing context. For example, advertisers may deliver targeted personalized advertisements which is likely to increase customer loyalty, and sales~\cite{chen2012personalized}. 

As the usage of smartphones and wearable technology continues to grow, inferring user behavior and context from noisy and complex sensor data becomes a difficult and challenging task. Various sensors can determine specific contexts. For example, accelerometer can be used for characterizing user's movements~\cite{lu2009soundsense} and microphone can detect environmental sound events~\cite{santos2010providing}. \cite{kwapisz2011activity} proposed a mobile context sharing system that uses five sensors (i.e., GPS, call logs, SMS, Bluetooth and battery) in order to infer user's activities. A single sensor may produce dozens of features, for example, accelerometer features are: size, x, y, and z axes values, correlations, and more. Thus, incorporating context into a CARS model increases both the dimensionality and sparsity of the model, making it a challenging problem~\cite{adomavicius2011context}. 

Another challenging aspect of RS and CARS, is how to constitute a useful system from the user perspective~\cite{sinha2002role}. Explaining recommendations to users increase the acceptance of the recommendations and gain the users' trust~\cite{cramer2008effects}. The complexity of recommendation algorithms often prevents users from comprehending recommended results and can lead to trust issues when recommendations fail. This objective is particularly important in RS that consider relevant contextual information such as CARS~\cite{verbert2012context}. Other user aspects are: privacy (e.g., using GPS instead of Wi-Fi to determine location may raise privacy issues), battery usage (e.g., maximize battery life in mobile-sensing applications~\cite{ben2009less}), etc. 

As we will show, the dimensionality challenge and the user-aspects challenge are related, since reducing the dimensionality may affect the user aspects. In order to address the dimensionality problem, the context representation must be reduced. Baltrunas et al.~\cite{baltrunas2011matrix} and others~\cite{consolvo2008activity, savage2012m,sun2013we} reduced the dimensionality by using domain knowledge to model situations and circumstances as explicit contexts. For example. weather conditions (sunny, rainy, etc.) or precise locations (home, work, etc.). While in this setting, the set of contexts is both small enough to handle and sufficient to prevent sparsity, it may not take into consideration other critical environmental features, and does not necessarily represent an optimal set of features for the recommendation process. Moreover, it only integrates a limited number of pre-defined explicit contextual information. Thus, the recommendations may fail and even lead to trust issues.

\begin{figure}[b]
    \centering
    \includegraphics[width=\linewidth]{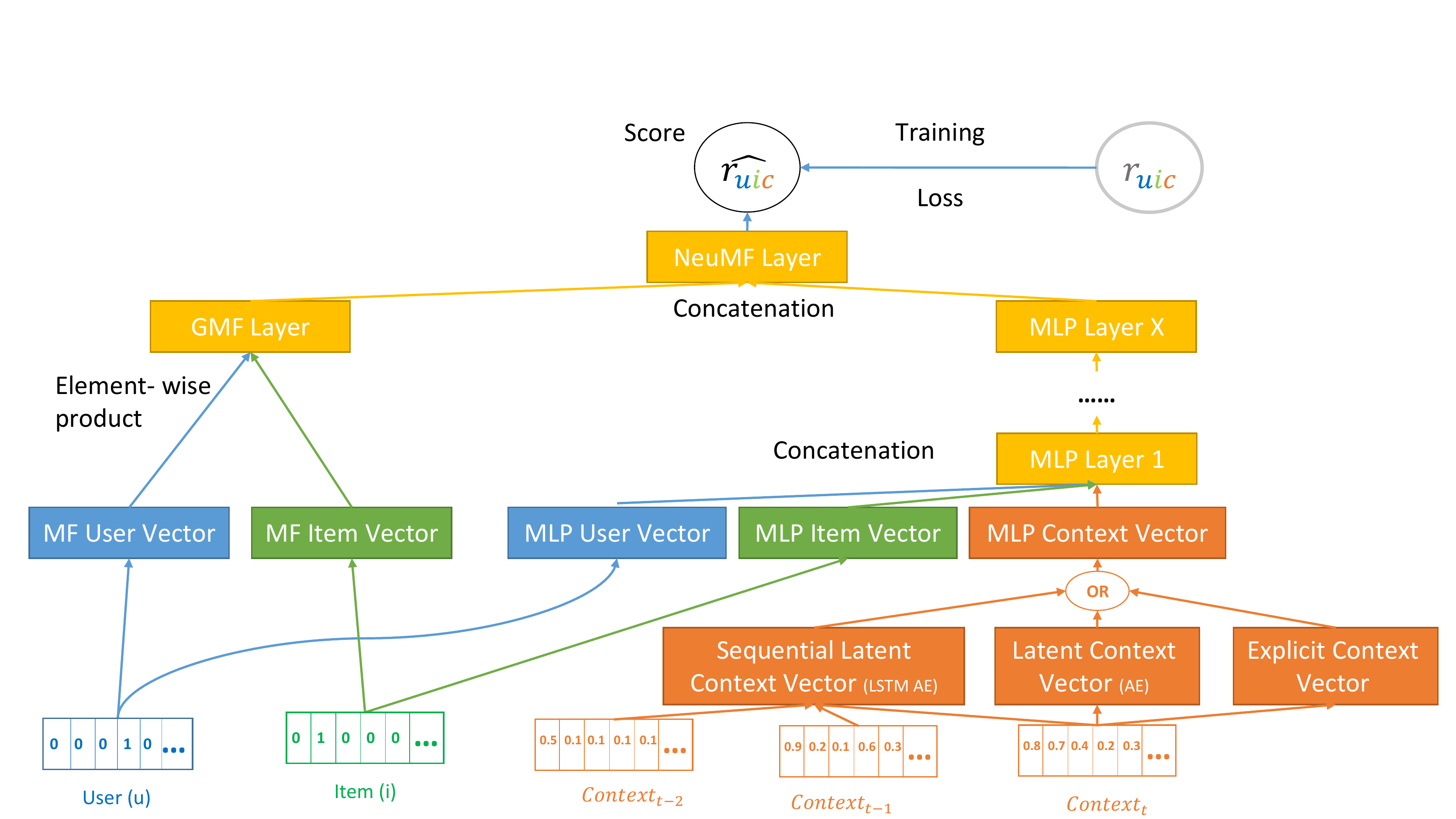}
    \caption{A deep contextual modeling architecture suggested by \cite{livne2019deep}}
    \label{fig:livne2019}
\end{figure}

Another approach, is to transfer the contextual information into a denser space. Livne et al.~\cite{livne2019deep} proposed to add a context component to \textit{NeuMF}~\cite{he2017neural} (also depicted in~\autoref{fig:livne2019}). Their representation allows for non-linear interactions between user, item, and context. Notably, they suggested an approach for incorporating high-dimensional contextual information in an explicit or a latent manner. Adding latent context to the recommendation process, improves the recommendation accuracy~\cite{livne2019deep, unger2018inferring} compared to~\cite{baltrunas2011matrix}. While the latent representation addresses the sparsity and dimensionality challenges within the contextual information, it suffers from several disadvantages. First, to transform into a denser latent space, all contextual features build on sensors must be sensed, and therefore it increases battery consumption. Second, sensing all sensors can sabotage users' aspects, such as privacy and trust. Finally, transferring into a latent space limits the ability to provide an explanation regarding the features that led to the recommendation.

% In~\cite{unger2016towards}, environmental features were represented as low-dimensional unsupervised latent contexts, extracted by an auto-encoder or by principal component analysis (PCA). 

In this paper we present an approach for selecting low-dimensional subsets of the contextual information and incorporating them explicitly within CARS. Specifically, we build on~\cite{livne2019deep} and present a novel feature-selection algorithm, based on \emph{genetic algorithms} (GA), that outperforms state-of-the-art dimensional-reduction CARS algorithms.

\newpage Our main contributions are summarized as follows:
\begin{enumerate}
    \item We present a GA-based feature-selection algorithm that outperforms state-of-the-art dimensional-reduction CARS algorithms in terms of accuracy.
    \item We develop two heuristics for a quick estimation of subsets' quality.
    \item We present two extensions to our baseline algorithm, that dramatically decrease the number of necessary sensors, while maintaining the high accuracy recommendations.
    \item Each of our extensions exploit GA traits for improving the explainability, and controlling user aspects, such as privacy and battery consumption.
    \item We provide an empirical analysis of our results on two high-dimensional context-aware datasets driven from smartphones.
    \item We provide many use cases and examples of how researchers and domain experts can tweak the feature-selection algorithm for improving the user aspects and the explainability.
\end{enumerate}

The remainder of the paper is organized as follows: In \autoref{sec:RelatedWork} we review related work. Our proposed methodology and our comprehensive evaluation are interwoven together in~\autoref{sec:method}. Finally, in~\autoref{section:Conclusions} we provide concluding remarks and discuss future work.

\section{Related Work} \label{sec:RelatedWork}
\subsection{Context-Aware Recommender Systems (CARS)}
There are several paradigms for incorporating contextual information in CARS. Adomavicius and Tuzhilin~\cite{adomavicius2011context} suggested three main approaches, namely pre-filtering, post-filtering, and contextual modeling. 

In both pre-filtering and post-filtering, the problem is reduced to the 2D setting, and the context is used during pre-processing or post-processing. Since the context is not fully integrated with the recommendation algorithm, exploiting the full potential of the relationships between various user-item combinations and contextual values, is prevented. 

The \emph{contextual-modeling} approach has been designed to explore this possibility, by using the contextual information directly in the recommendation function. As mentioned before, this approach is challenging due to the dimensionality expansion and the sparsity of the model. In~\autoref{sec:intro}, we described the explicit approach~\cite{baltrunas2011matrix, consolvo2008activity, savage2012m, sun2013we, zheng2014cslim}, that model situations and circumstances as explicit specific contexts. We also described the latent approach~\cite{unger2016towards, unger2018inferring, xin2019cfm, wu2017improving}, that add latent context to the recommendation process, e.g., by represent environmental features as low-dimensional unsupervised latent contexts, extracted by an auto-encoder (AE) or by principal component analysis (PCA)~\cite{unger2016towards}. While the latent representation addresses the sparsity and static challenges within the contextual information, it considers only linear interactions. In the following paragraph we talk about additional latent approaches.

Recent studies suggest fitting the rating data using a regression model, such as tensor factorization (TF)~\cite{karatzoglou2010multiverse} that extends the classic two-dimensional matrix-factorization problem to an n-dimensional version. Simple CARS based on a matrix factorization (MF) model with a reduced set of parameters (linear with the number of contexts), may obtain equal or improved recommendations results~\cite{baltrunas2011matrix}. The challenge however, is that the number of model parameters grows exponentially with the number of contextual factors. 
 
\textit{NeuMF}~\cite{he2017neural} mimics MF using deep learning, a method with a rising popularity when used with RS~\cite{zhou2015priori}. Several deep models have been recently suggested for CARS as well. \cite{kim2016convolutional} learned the item representation by combining a convolutional neural-network (CNN) model with MF. Their method considers only one type of item feature as contextual information. \cite{xin2019cfm} proposed utilizing convolutional neural networks (CNNs) for enhancing a factorization-machines (FM) algorithm, named CFM, that models high-order interactions between contextual variables. \cite{costa2019collective} proposed to model the user, item, and time embeddings to capture temporal patterns. However, in this work, only a single, predefined, explicit contextual information, was modeled.

\subsection{Genetic Algorithms (GA)}
Genetic algorithm (GA) is a search heuristic, inspired by the Darwin's natural evolution~\cite{eiben2003introduction}. A GA takes a population of candidate solutions to an optimization problem, (i.e., individuals, or phenotypes), and transforms them through an iterative process (i.e., evolution) where at each iteration (i.e., generation), the individuals are: 1) evaluated for their performance using an objective function (i.e., fitness function); 2) stochastically selected according to fitness; and 3) modified by means of recombination (i.e., crossover) and mutation. The modification operators act on the individuals' genetic representation (i.e., genome), that is usually an array of bits, numbers, or other types. The algorithm usually terminates when reaching a satisfactory fitness level, or when a maximum number of generations have passed. 

In this paper we utilize GA for selecting contextual features for CARS. To the best of our knowledge, GA in particular and evolutionary computation techniques in general, have never been applied to the domain of CARS, though there much work has been done on utilizing GA for feature selection (FS)~\cite{liu2008wrapper, xue2015survey, liu2010feature}. Yet, applying GA-based FS for neural networks (NN), is a non-trivial task, since the fitness evaluation of each individual at each generation (typically thousands of evaluations), requires, at least in the na\"ive approach, to train of the NN and estimate its accuracy. 

To overcome this problem, \cite{jiang2017modified} used a pre-trained deep neural network (DNN) restricted in the number of layers and nodes in each layer (3 and 20, respectively) for evaluation. The individuals fitness was calculated by predicting the DNN using the individuals. At the end of the run, a subset of the entire population is selected and stacked into a latent representation that compresses the genetic material of the selected individuals. It should be noted that the authors validated their method on a small dataset with only 60 features, while in this paper we deal with two datasets with 480 and 661 features. Nevertheless, we have adopted some of the principle ideas of~\cite{jiang2017modified}, though we adjusted them to our domain and took a different approach for implementing them. For example, the restriction on the number of nodes forced the authors to translate the features into a latent representation, which we wanted to avoid. Moreover, the authors presented complex operators for initializing the population and for recombining and mutating the individuals. We preserved the simple and effective standard GA operators while maintaining high accuracy (as elaborated in~\autoref{subsec:basic_solution}). We also adopted the idea of exploiting the knowledge that was learned by the multiple generated individuals and stacking them together, though we applied it differently and not on the individuals (as elaborated in~\autoref{sec:method:ensemble}).

% https://www.mdpi.com/1996-1073/11/7/1636
% https://www.sciencedirect.com/science/article/pii/S0950705117303209?casa_token=lkyEpblvTXgAAAAA:iHlVt9hd97TFluqKoQep6BiSr-tzqNNJxE5ftOuv_XBlG3XauZWQ1MIGJepMgF886N9OGVs

\section{Method}
\label{sec:method}
Our method uses the architecture proposed by~\cite{livne2019deep,unger2020context} (depicted in~\autoref{fig:livne2019}) with four major changes: (1) the architecture was originally designed to solve a regression task (i.e., rating prediction), while here we use it for solving binary classification problems; (2) we replace the mechanism for creating the MLP Context Vector with a GA-based mechanism (elaborated in~\autoref{subsec:basic_solution}); (3) we further refine the GA algorithm with a better control over the selected features and contextual dimensions (elaborated in~\autoref{sec:sensing_extentsion}); and (4) we add an ensemble step to the architecture of~\cite{livne2019deep}, exploiting unique GA traits for improving the explainability, and for controlling user aspects, such as privacy and battery consumption.

We note that we call change 2 a \emph{basic solution}, since it is ``self contained'', meaning that it outperforms all the baselines in several aspects (as described below). To keep the flow of the paper, we conclude the description of each change, with its evaluation.

\subsection{Problem Formulation}
\label{sec:method:problem_formulation}
The input for the proposed architecture (all steps) is a set $D$ composed of $N$ tuples. Each tuple $(u, i, c, Y)\in D$ denotes an interaction event where user $u$ was exposed to an item $i$, while considering contextual information $c$ regarding this interaction. $c$ 
% includes additional information about $u$ and $i$ 
may have two representations: categorical (e.g., weather condition, etc.), and continuous fields (e.g., orientation, accelerometer, etc.). Finally, $Y\in\{0,1\}$ is the associated label of the interaction event, indicating the user rated feedback, where $Y=1$ indicates that user $u$ provided a positive rating for item $i$, under contextual information $c$, and $Y=0$ indicates a negative rating.

\subsection{Datasets}
\label{sec:method:datasets}
We use two high-dimensional, context-aware datasets for evaluation, both were used in previous papers and has been provided to us by their authors. We note that we focused on these two datasets since the number of available datasets with rich contextual data is scarce. The contextual features were extracted from multiple contextual dimensions, such as accelerometer, gravity, etc'. For each contextual dimension, various statistical contextual features were extracted, such as average, standard deviation, entropy, etc. The datasets are (also summarize their features in \autoref{tab:datasetStats}): 

\begin{itemize}
    \item \textbf{CARS~\cite{livne2019deep,unger2018inferring,unger2020context,unger2019hierarchical}:} Contains 38,900 explicit ratings (dislike, like and check-in) of 1,918 points of interest (POIs), each rating being associated with 480 contextual features. The data was collected from various types of contextual dimensions, including environmental information, user activities, mobile state, and user behavioral data. To transform the ratings into a binary scale, we converted `like' and `check-in' to one, and `dislike' to zero.
    \item \textbf{Hearo~\cite{unger2016towards,unger2019hierarchical}:} Derived from a field experiment in which users interacted with a recommender system that provided recommendations of POIs and received users' binary ratings on the provided recommendations. The ratings have been obtained by 77 users and it associates with 661 contextual features. 
\end{itemize}

We split each dataset as follows: the last 20\% of the interactions were considered as a test set. Of the remaining 80\%, the last 10\% were used as a validation set and the rest 70\% were used for training. 

\begin{table}
\centering
\begin{tabularx}{\textwidth}{>{\bfseries}lCC}
\toprule
    & \textbf{CARS Dataset}& \textbf{Hearo Dataset}
\\
\midrule
\# users & 98 & 77  \\
\# items & 1,918 & 228 \\
\# ratings & 38,900 & 7,416\\
rating scale & $0=\text{dislike}, 1=\text{like and check-in}$ & 0-1\\
rating sparsity & 96.41 & 57.75 \\
\# contextual dimensions & 14 & 16 \\
\# contextual features & 480 & 661\\ \midrule
% \textbf{contextual dimensions} & \small{time, weather, ringer mode, accelerometer, orientation, location, running applications, screen log, battery, light, network traffic, gravity, microphone, magnetic field}  & \small{time, location, ringer mode, battery, activity recognition, light, accelerometer, orientation, application traffic, gravity, microphone, screen log, weather, magnetic field, cell state}
\# contextual dimensions 
& GPS (2), Weather (2), Activity (8), Mic (12), Light (12), Orientation (71), Magnetic (71), Accelerometer (71), Gravity (71), Internet related (102),
Calls (21), Data received and sent (4), Application Related (24), Other (9)
& Weather (21), GPS (2), Mic (13), Light (21), Activity (6), Orientation (71), Magnetic (71), Accelerometer (71), Rotation (71), Gravity (71), Gyroscope (71), Internet related (102), Calls \& SMS (44),
Application related (17), Time (6), Other (3)\\\hline
\end{tabularx}
\vspace{1ex}
\caption{The context-aware datasets we used for evaluation}
\label{tab:datasetStats}
\end{table}

\subsection{Evaluation Metrics}
We used the following evaluation metrics in our experiments, both are widely used for evaluation of binary-classification predictions~\cite{guo2017deepfm, huang2019fibinet, liu2019feature}:

\begin{itemize}
    \item \textbf{AUC} --- the area under the ROC curve. The metric is insensitive to the classification threshold and the positive ratio. The upper bound of the AUC is one, and the higher the AUC, the better.
    \item \textbf{Log loss} --- measures the distance between two distributions. The lower bound of the log loss is zero, indicating that the two distributions match perfectly, and a lower values indicates better performance.
\end{itemize}

\subsection{Baselines}
\label{subsec:baselines}
Using the evaluation metrics, we compared our method to the following baselines (for each of the changes):

\begin{enumerate}
    \item \textbf{Model with no contextual features:}
    \begin{itemize}
        \item Neural Matrix Factorization (\emph{NeuMF})~\cite{he2017neural}  --- mimics MF using deep learning without using any contextual information %. elaborated in~\autoref{sec:RelatedWork} .
    \end{itemize}
    
    \item \textbf{Deep context-aware models with feature selection:}
    \begin{itemize}
        \item Explicit neural context model (\textit{ENCM})~\cite{livne2019deep} --- deep context-aware model utilizing explicit context conditions. For explicit context features, we followed \cite{baltrunas2011matrix} and chose the time of day, time of week, and weather. %elaborated in~\autoref{sec:intro}.
        % - 
        \item Forward selection context model (\emph{FSCM})~\cite{chandrashekar2014survey} --- utilizes the features selected by the sequential feature selection (SFS) and incorporates them as explicit context within ENCM. SFS is an iterative method in which we start with having no feature in the model. In each iteration, we keep adding the feature which best improves our model till an addition of a new variable does not improve the performance of the model.
    \item Backward selection context model (\emph{BSCM})~\cite{chandrashekar2014survey} --- like FSCM, but uses the sequential backward selection (SBS).
    In SBS, we start with all the features and removes the least significant feature at each iteration which improves the performance of the model. We repeat this until no improvement is observed on removal of features.
    \end{itemize}
    
    \item \textbf{Deep context-aware models with dimensionallity reduction:}
    \begin{itemize}
        \item Latent neural context model (\textit{LNCM})~\cite{livne2019deep} --- deep context-aware model which considers low-dimensional latent context via auto-encoder network.
        % elaborated in~\autoref{sec:intro}.
        % 
        \item Sequential neural context model (\textit{SLCM})~\cite{livne2019deep} --- deep context-aware model which considers low-dimensional sequential latent contexts extracted via encoder-decoder LSTM network.
    \end{itemize}
    
    \item \textbf{Deep context-aware models without feature reduction nor selection:}
    \begin{itemize}
        \item Neural context model (\textit{NCM})~\cite{livne2019deep} --- incorporates all contextual information within ENCM as explicit context.
        \item Deep factorization machines (\textit{DeepFM})~\cite{guo2017deepfm} --- integrates the architectures of FM and DNNs, by modeling low-order feature interactions as FM and modeling high order feature interactions as DNNs.
        \item Combining feature importance and bi-linear feature interaction (\textit{FiBiNET})~\cite{huang2019fibinet} --- a DNN model for calculating the feature interactions using a bi-linear function.
    \end{itemize}
\end{enumerate}

DeepFM and FiBiNET baselines are state-of-the-art algorithms for solving binary classification problems like CTR prediction. Thus, we included them among our baselines. In order to implement them we used the implementation of \textit{DeepCTR}\footnote{DeepCTR: \url{https://github.com/shenweichen/DeepCTR}}. We implement the remain baselines using Keras~\cite{chollet2015keras}. Our Code is available at \url{https://github.com/recsysGA-CARS/GA_CARS_2020}.

% \subsection{Class Distributions}
% \label{sec:class_dist}
% The first component of our method is representing the contextual information with GA utilizing the sub-network MLP. For this component, we weighted each class in order to achieve a better representation of the contextual information corresponding to the labeled data.
% We conduct our experiments on two datasets, CARS and Hearo (as described in \autoref{subsec:datasets}). While the Hearo dataset label data is balanced, the CARS dataset label data is imbalanced .To deal with this imbalance ratio and treat both classes equally, we set a weight for each class; in this way, we can mimic a balanced distribution. The weights are determined by the distribution of the training set as follows: 

% \begin{equation}
% Class Weights= 
% \begin{cases}
% $not relevant recommendations$ & 1.0\\
% $relevant recommendations$ & \frac{Count(training-set_{not-relevant-recommendation})}{Count(training-set_{relevant-recommendation})} > 1.0
% \end{cases}
% \end{equation}

\subsection{Parameters}
When using the deep contextual modeling architecture of~\cite{livne2019deep}, we set the dimensions of all embedding layers to 16, the batch size to 256, and applied a dropout rate of 50\% for the NeuMF layer. For optimization, we used the Adam optimizer~\cite{kingma2014adam} and stochastic gradient descent (SGD), as described in~\cite{he2017neural}. For computing the loss we used binary cross entropy. Furthermore, we conducted the experiments using RTX 2080 TI GPUs.

% \section{Method}
% \label{section:Method}
% % We implemented GA-CARS using DEAP\footnote{DEAP: https://github.com/deap/deap} \cite{DEAP_JMLR2012} and Keras\footnote{Keras: https://keras.io/} \cite{chollet2015keras}.
% % We compare the performance of GA-CARS \footnote{Codes are available at https://github.com/eliadsbgu/GA\textunderscore{CARS}} with the following baselines:

% In this section we lay out three different solutions for the above problem. 
% For simplicity, we begin with a basic solution, where all of the following extension are based on top of this solution. In our first extension solution we suggest to edit the evaluation function in order to give an incentive for each individual to reduce the number of context dimensions used rather then reduce the number of contextual features only. In the last extension, we suggest to take advantage of GA multi solution oriented properties in order to produce an ensemble model. The ensemble model is build upon several different individual. 

% We believe that our suggested solutions are providing a novel overlook of incorporating contextual information within CARS. Moreover, our solutions emphasize users' aspects in terms of battery consumption, privacy, and transparency in addition to preserving a high recommendation accuracy rate. Thus, they all provide an important added value. 

\subsection{Basic Solution}
\label{subsec:basic_solution}
The basic solution replaces the mechanism of~\cite{livne2019deep} for creating the MLP Context Vector, with an evolutionary-based mechanism. For this end, we devised a GA-based feature-selection (FS) algorithm, where each individual represents a subset of the contextual information. The subsets are improved over the course of evolution, in terms of accuracy and size. Once the evolutionary process ends, the best individual is chosen for deciding which features will be explicitly included during the learning phase of the complete architecture of~\cite{livne2019deep}. We used DEAP~\cite{DEAP_JMLR2012} for the implementation of the GA algorithm and the evolutionary operators. Below we describe the elements of our setup in detail.

\paragraph{Genome.} The natural representation for each individual is a binary string of the length of the number of contextual features to consider, representing the absence or presence of each of the contextual features. This representation allows for using standard GA operators (i.e., binary crossover and binary mutation). We initialized the experiment with a random population, where the probability for a value of one was sampled from a normal distribution with $\mu=0.25 \times \mathtt{\#\,of\,contextual\,features}$ and $\sigma=40$.

\paragraph{GA Operators and Parameters.} We used a standard k-tournament selection with $k=5$, and bitwise operators --- 5-point crossover, n-point crossover and n-point mutation~\cite{eiben2003introduction}. We experimented with several parameter settings, finally settling on: population size --- 100, generation count --- 300, crossover probability --- 0.65 (where 5-point crossover and n-point crossover were selected with the same probability), and probability for flipping a bit (i.e., mutation) --- 0.025. We used a uniform distribution for selecting crossover points within individuals. 

% A GA contextual information representation is generated for each individual. The inputs for this step are two vector with the same length: contextual information vector denoted by $c$ and an individual binary vector denoted by $i$. Every bit $b$ ,$b\in\{0,1\}$, in $i$ is  representing whether to consider the corresponding contextual feature in $c$ or not. We apply an element wise multiplication between $c$ and $i$ and generate a GA contextual representation denoted by $ci$. We apply an element wise multiplication between $c$ and $i$ and generate a GA contextual representation denoted by $ci$. 

\begin{figure}
    \centering
    \includegraphics[width=\textwidth]{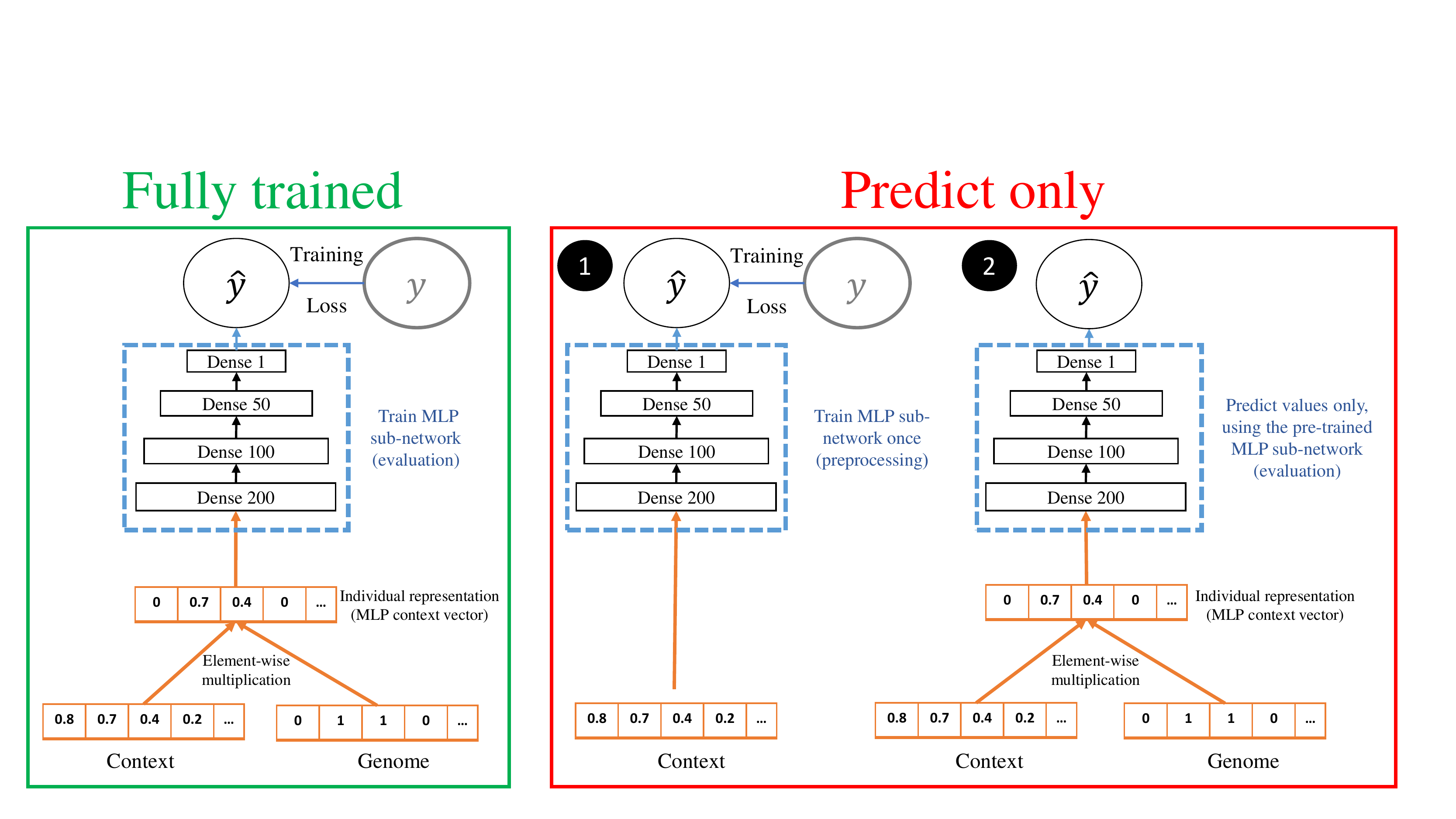}
    % \vspace{-2em}
    \caption{Generating GA Contextual Solutions Methods}
    \label{fig:GA_Representations}
\end{figure}

\paragraph{Fitness.} A major problem we had to tackle is the long time required for evaluating the individuals. A na\"ive approach would be to train the deep contextual modeling for each individual, resulting with a total of 30K calculation (300 generations $\times$ 100 individuals). Since each calculation takes approximately one minute, this is not feasible. Thus, inspired by the work of~\cite{jiang2017modified} (see~\autoref{sec:RelatedWork}), we devised two heuristic approaches for a quick estimation of the individuals AUC when used for training the deep contextual model (depicted in~\autoref{fig:GA_Representations}):
% Notably, for this step, we use the contextual features only, without any information regarding the user or item. Then we train a single deep contextual modeling build upon a the best individual in terms of accuracy as follows:

\begin{enumerate}
    \item \emph{Fully trained} --- Use the individual under evaluation for training a multi-layer perception (MLP) sub-network. This operation takes approximately five seconds per individual.
    \item \emph{Predict Only} --- This heuristic includes a pre-processing step to the evolutionary process, where we train once a robust single model (an MLP sub-network), using all available contextual information. Notably, the bias component of each layer is disabled, relying on the weights only, in order to prevent the bias from influencing our evaluation score. During evolution individuals are estimated by predicting the output of the pre-trained network with using the individual's chosen contextual features. This operation takes approximately 60ms and the pre-processing takes approximately five seconds.
\end{enumerate}

In addition to the AUC estimation of the individuals, we wanted to reinforce individuals that represent small-sized subsets. The motivation for this is related to the user aspects that we discussed in~\autoref{sec:intro} (i.e., privacy, battery consumption and transparency). Reducing the size of the subsets while maintaining a high accuracy, addresses some user aspects, such as privacy and battery consumption. Also, since the transparency is addressed by the individuals representation of explicit MLP context vectors. 

Thus, we ended up with the following fitness function for an individual $i$: 

$$f_i =  W_{AUC}\times\frac{1-AUC_i}{1-AUC_{min}} + W_{size} \times \frac{size_i}{size_{max}},$$ where $W_{AUC}=0.8$ and $W_{size}=0.2$ are the weights of the AUC and the size components (respectively), $AUC_i$ and $size_i$ are the estimated accuracy and the size of the subset of individual $i$ (respectively), and $AUC_{min}$ and $size_{max}$ are the minimal/maximal AUC/size in the current generation (respectively), used for normalizing the two components.

% We use $ci$ as an input to a MLP network in order to predict the binary target label $Y$. This component results with a multitude of solutions $N$. Each solution $s\in N$ is determined by $i$ and $c$. We denote the set of features that are equal to one in the corresponding individual $i$ as $s_{features}$. we evaluate all $N$ solutions in terms of recommendation accuracy (i.e. AUC).

% While recommendation accuracy is an important goal in CARS, other user aspects (i.e., privacy, battery consumption and transparency) should be considered and cannot be ignored. However, most current studies in the CARS domain focus mainly on the accuracy of the recommendations and neglect the users' aspects. To address this gap, we suggest to estimate (i.e., fitness function) each individual based on its recommendation accuracy (i.e., AUC score) and number of features used within the individual (i.e., $s_{features}$). Thus, we suggest giving each individual an incentive to minimize the number of features incorporating in the deep contextual model and remain in the explicit space for transparency reasons.

% We evaluated each individual' quality by the following equations: 
% \begin{equation}
% \label{eq:basic_solution_fitness}
%     fitness(individual) =  W_{AUC}\times\frac{1-AUC(s)}{1-Min(AUC\,in\,generation)}+ W_{features}\times\frac{len(s_{features})}{Max(len(s_{features}) in generation)})
% \end{equation}

% We train a \textbf{single} deep contextual model based on the best individual in terms of AUC utilizing the sub-network MLP for both Fully-Train and Only-Pred approaches.

\subsection{Basic Solution - Results}
We compared the basic solution to state-of-the-art approaches for CARS, i.e., the baseline algorithms (see~\autoref{subsec:baselines}). Tables~\ref{tab:basic_solution_cars} and~\ref{tab:basic_solution_Hearo} summarize the findings for the CARS and the Hearo datasets (respectively). For each algorithm, we stated the number of contextual dimensions it used, the number of features, whether the output is explicit, and finally, the AUC and log loss metrics. The data for the basic solution has been derived from the best individual of one of the runs, though other runs exhibited similar results.

% \begin{table}[h]
% \caption{Comparison of Baselines to our Basic Solution - CARS Dataset}
%  \label{tab:basic_solution_cars}
% \begin{tabular}{ll|c|c|c|c|c|c}

%  &  & \begin{tabular}[c]{@{}c@{}}\# Contextual \\ Features Used\end{tabular} & \begin{tabular}[c]{@{}c@{}}\# Contextual\\  Dimensions Used\end{tabular}  & AUC (\%) & Log loss & \begin{tabular}[c]{@{}c@{}} Transparency \\ (V/X) \end{tabular} \\ \hline
% No Contextual Features Models & NeuMF & 0 & 0 &  79.53 & 0.481 & V     \\ \hline
% \begin{tabular}[c]{@{}l@{}}Deep Context Aware Models\\  With Feature Selection\end{tabular} & ENCM & 6 & 2 &  79.25 & 0.482 & V    \\
%  & FSCM & 9 & 5 &  79.89 & 0.459 &    V\\
%  & BSCM & 478 & 14 &  80.56 & 0.448 &  V \\ \hline
% \begin{tabular}[c]{@{}l@{}}Deep Context Aware Models\\ With Feature Reduction\end{tabular} & LNCM & 480 & 14 &  79.93 & 0.454 &     X \\
%  & SLCM (Lookback =3) & 480 & 14 &  80.04 & 0.454 &      X\\
%  \hline
% \begin{tabular}[c]{@{}l@{}}Deep Context Aware Models \\ Without Feature Reduction or Selection\end{tabular} & NCM & 480 & 14 &  80.24 & 0.451 &  V    \\
%  & DeepFM & 480 & 14 &  75.95 & 0.499 &      V\\
%  & FiBiNET & 480 & 14 &  76.23 & 0.517 &    V\\ \bottomrule
% Basic Solution & Full-Train & 88 & 12 &  \textbf{80.94} & \textbf{0.443} &    V\\
%  & Only-Pred & 51 & 11 &  79.97 & 0.460 &    V\\ \hline
% \end{tabular}
% \end{table}

\begin{table}[b]
\centering
\begin{tabular}{ll|>{\centering}p{1.7cm}|>{\centering}p{1.7cm}|>{\centering}p{1.1cm}|c|c}
Type & Algorithm & \# Contextual Dimensions & \# Contextual Features & Explicit? (V/X) &AUC & Log loss  \tabularnewline \hline
No contextual features & NeuMF & 0 & 0 & V& 0.7953 & 0.481  \tabularnewline \hline
Deep context-aware & ENCM & 2 & 6 & V& 0.7925 & 0.482  \tabularnewline
w/ feature selection & FSCM & 5 & 9 &V& 0.7989 & 0.459  \tabularnewline
 & BSCM & 14 & 478 & V& 0.8056 & 0.448  \tabularnewline \hline
Deep context-aware & LNCM & 14 & 480 & X& 0.7993 & 0.454  \tabularnewline
w/ feature reduction & SLCM & 14 & 480 & X&0.8004 & 0.454  \tabularnewline \hline
Deep context-aware & NCM & 14 & 480 & V& 0.8024 & 0.451 \tabularnewline
w/o dimensionality & DeepFM & 14 & 480 & V& 0.7595 & 0.499  \tabularnewline
reduction & FiBiNET & 14 & 480 & V&0.7623 & 0.517  \tabularnewline \hline
Basic solution & Fully trained & 12 & 127 & V& \textbf{0.8108} & \textbf{0.442}  \tabularnewline
 & Predict only & 12 & 65 & V& 0.8067 & 0.446  \tabularnewline \hline
\end{tabular}
\vspace{1ex}
\caption{Our basic solution compared to the baselines --- CARS Dataset. Our solution outperforms all baselines.}
\label{tab:basic_solution_cars}
\end{table}

\begin{table}[ht]
\centering
\begin{tabular}{ll|>{\centering}p{1.7cm}|>{\centering}p{1.7cm}|>{\centering}p{1.1cm}|c|c}
Type & Algorithm  & \# Contextual Dimensions & \# Contextual Features & Explicit? (V/X) & AUC & Log loss \tabularnewline \hline
No contextual features & NeuMF & 0 & 0 & V& 0.8007 & 0.542  \tabularnewline \hline
Deep context-aware & ENCM & 2 & 17 & V& 0.8177 & 0.556  \tabularnewline
w/ feature selection & FSCM & 5 & 6 & V& 0.8164 & 0.529  \tabularnewline
 & BSCM & 16 & 658 & V& 0.8160 & 0.526  \tabularnewline \hline
Deep context-aware & LNCM & 16 & 661 & X& 0.8095 & 0.532 \tabularnewline
w/ feature reduction & SLCM & 16 & 661 & X& 0.8203 & 0.518  \tabularnewline \hline
Deep context-aware & NCM & 16 & 661 & V& 0.8191 & 0.519 \tabularnewline
w/o dimensionality & DeepFM  & 16 & 661  & V& 0.7178 & 0.652  \tabularnewline
reduction & FiBiNET & 16 & 661  & V & 0.7294 & 0.646  \tabularnewline \hline
Basic solution & Fully trained & 16 & 96 & V& 0.8162 & 0.520  \tabularnewline
 & Predict only & 15 & 136 & V&\textbf{0.8228} & \textbf{0.516}  \tabularnewline \hline
\end{tabular}
\vspace{1ex}
\caption{Our basic solution compared to the baselines --- Hearo Dataset. Our solution outperforms all baselines.}
\label{tab:basic_solution_Hearo}
\end{table}

As presented in the tables, our approach outperform all the baselines in both AUC and Log Loss metrics, while using explicit features only, and addressing user aspects. While the results are interesting, we believe that our method has other important advantages that lies within the tables and the raw results data. 

For example, the baseline with the highest AUC and log loss on the CARS dataset --- BSCM, used 478 contextual features and relies on all contextual dimensions. On the other, our best individual used all contextual dimensions, except for GPS and network traffic, thus, addressing user aspects as privacy and battery consumption. In the Hearo dataset, the best baseline -- SCLM, utilizes a latent space, therefore it used all 661 contextual features and all contextual dimensions, while out best individual used solely 136 contextual and 15 contextual dimensions. Moreover, our individual did not used the users activities information, again, addressing user aspects.

\begin{figure}
    \centering
    \includegraphics[width=\textwidth]{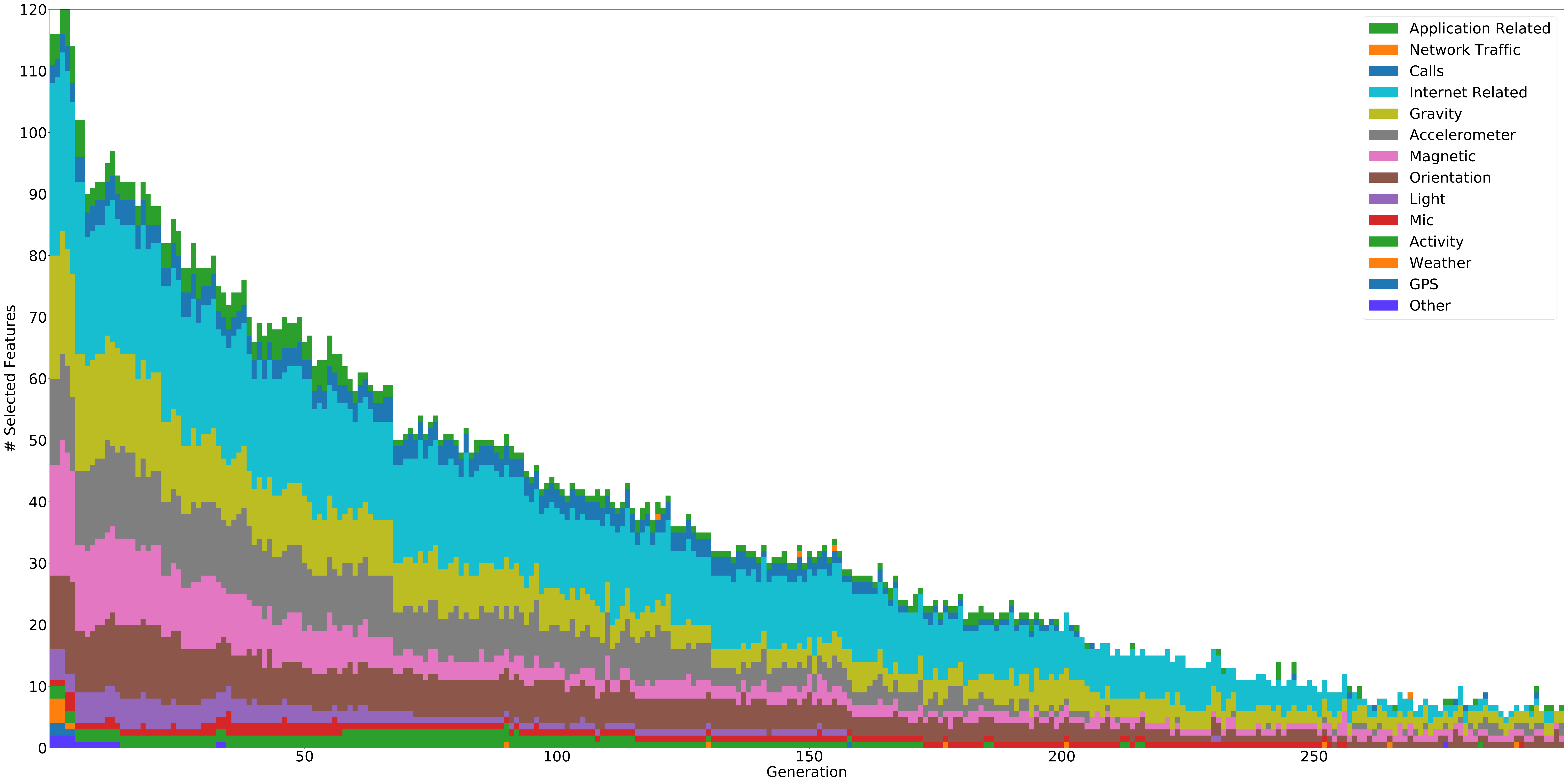}
    \caption{CARS best individuals. For each generation the graph shows the number of features that the best individual used, grouped by dimensions.}
    \label{fig:basic_graph}
\end{figure}

\autoref{fig:basic_graph} presents the number of features that the best individual used at each generation, grouped by dimensions. Interestingly, the ratio between the different contextual dimensions is kept most of the time, suggesting that if we wish to further address the user aspects and eliminate dimensions, we will need to change the evolutionary process, as we will show in the following section.

\subsection{Extension \#1 --- Minimizing Contextual Dimensions}
\label{sec:sensing_extentsion}
In this section, we wish to further improve the basic solution by reducing the number of contextual dimensions of the individuals. We propose to achieve that by adding a `dimensions' component to the fitness function. 

Our new fitness function is:
$$f_i =  
  W_{AUC}\times\frac{1-AUC_i}{1-AUC_{min}} + 
  W_{size} \times \frac{size_i}{size_{max}} + 
  W_{dim} \times \varphi \left( \frac{dim_i}{dim_{max}}\right),$$ 
where $W_{AUC}=0.8,W_{size}=0.15,W_{dim}=0.05$ are the weights of the AUC, the size, and the dimension components (respectively), $dim_i$ is the number of selected dimensions (sensors), $dim_{max}$ is the maximal $dim_i$ value in the current generation, and $\varphi(x)= \frac{1-e^{-x}}{1-e^{-1}}$ is a \emph{selection-pressure} function, that exponentially improves the component's score for each decrease in the number of used dimensions. The rest of the symbols remain the same (see~\autoref{subsec:basic_solution}). 

\subsection{Extension \#1 - Results}
In tables \ref{tab:family_solution_cars} and \ref{tab:family_solution_Hearo} we compare the performance of this extension to the performance of the basic solution and the best baseline. 

\begin{table}[ht]
\begin{tabular}{ll|>{\centering}p{1.7cm}|>{\centering}p{1.7cm}|>{\centering}p{1.1cm}|c|c}
Type & Algorithm & \# Contextual Dimensions & \# Contextual Features & Explicit? (V/X) & AUC & Log loss \tabularnewline \hline
Best baseline & BSCM & 14 & 478 & V& 0.8056 & 0.488  \\ \hline
Basic solution & Fully trained & 12 & 88 & V& \textbf{0.8094} & \textbf{0.443}  \\
 & Predict only & 11 & 51 & V& 0.7997 & 0.460  \\ \hline
Extension \#1 --- & Fully trained & \textbf{9} & 81 &V& 0.8075 & 0.445 \\
minimizing dimensions & Predict only & 5 & 34 & V& 0.7984 & 0.460 \\ \hline
\end{tabular}
\vspace{1ex}
\caption{The performance of the first extension compared to the basic solution and the best baseline --- CARS dataset.}
 \label{tab:family_solution_cars}
\end{table}

\begin{table}[ht]
\begin{tabular}{ll|>{\centering}p{1.7cm}|>{\centering}p{1.7cm}|>{\centering}p{1.1cm}|c|c}
Type & Algorithm & \# Contextual Dimensions & \# Contextual Features & Explicit? (V/X) & AUC & Log loss \tabularnewline \hline
Best baseline & SLCM & 16 & 661  & X& 0.8203 & 0.518   \\  \hline
Basic solution & Fully trained & 16 & 96  & V& 0.8162 & 0.520   \\
 &  Predict only & 15 & 136  & V&\textbf{0.8228} & \textbf{0.516}   \\ \hline
Extension \#1 --- & Fully trained & 9 & 94  &V& 0.8108 & 0.527  \\
minimizing dimensions &  Predict only & \textbf{8} & 123 & V& 0.8154 & 0.521  \\ \hline
\end{tabular}
\vspace{1ex}
\caption{The performance of the first extension compared to the basic solution and the best baseline --- Hearo dataset.}
 \label{tab:family_solution_Hearo}
\end{table}

Reducing the number of contextual dimensions resulted with less accurate results in terms of AUC and log loss. Yet, the change in accuracy is minor, compared to the change in the contextual dimensions. When comparing in CARS the AUC of the fully-trained individual of the basic-solution and the extension, we observe a decrease in the AUC of $0.0019$ from the basic solution, though the extension's individual used only 9 contextual dimension, compared to 12 of the basic extensions (and compared to the 14 of the best baseline). Moreover, the extension's individual did not use the following contextual dimensions: GPS, orientation, accelerometer and weather conditions. Thus, this extension provides improvements in both battery consumption and privacy compared to all other baselines. Similarly, in the Hearo dataset the change in the number of contextual dimensions was even more dramatic, reducing it in almost a half for the fully-trained individual. 

\begin{figure}
    \centering
    \includegraphics[width=\textwidth]{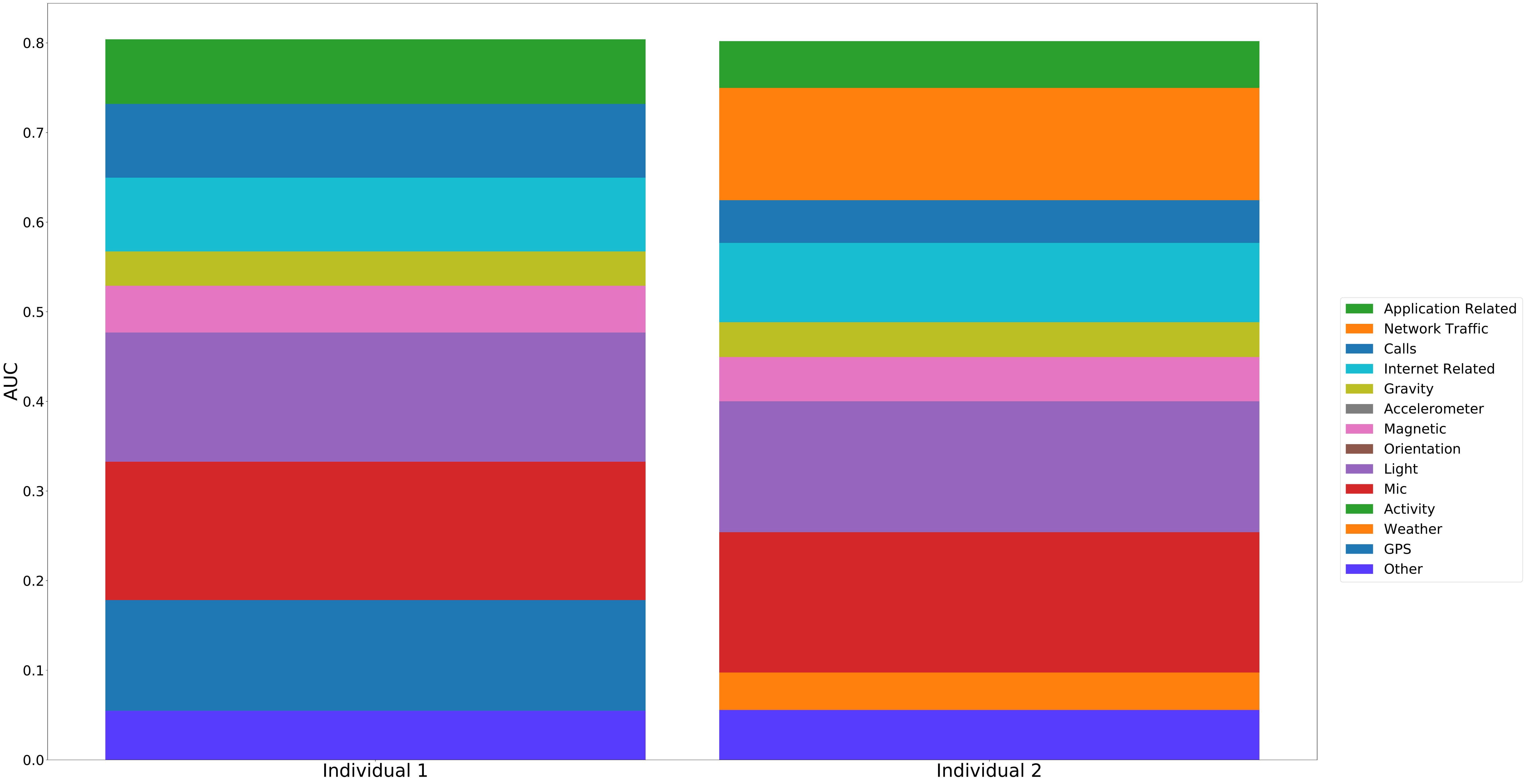}
    \caption{Comparing the contextual dimensions of two individuals of the same experiment with similar accuracy. While they share most contextual dimensions, the first individual relies on GPS while the other relies on weather and network traffic instead.}
    \label{fig:family_solution}
\end{figure}

Another interesting result is related to the evolutionary process, that generates a vast amount of individuals with a high accuracy (though not the highest). When we examined two different individuals \emph{of the same experiment} that yielded a similar AUC of 80.04 and 80.02, we observed that they used different contextual dimensions. As depicted in~\autoref{fig:family_solution}, while both individuals share most of the contextual dimensions, the first individual relies on GPS, while the second relies on weather and network traffic instead. Since both individuals performed similarly in terms of accuracy, the second individual can be used, for example, when the GPS sensor is undesired due to privacy and battery-consumption aspects.

To summarize these results, the `dimension' extension succeeded with its goal to further improve user aspects such as battery consumption and privacy. The question is --- can we do better? Can we improve these user aspects without compromising the accuracy?

\subsection{Extension \#2 --- Individuals Ensemble}
\label{sec:method:ensemble}
Considering the two individuals of~\autoref{fig:family_solution}, we hypothesized that an ensemble of different individuals with a similar accuracy will yield even better results. Such an approach will exploit the already generated individuals and will fulfill their potential. 

\begin{figure}[ht]
    \centering
    \includegraphics[width=\linewidth]{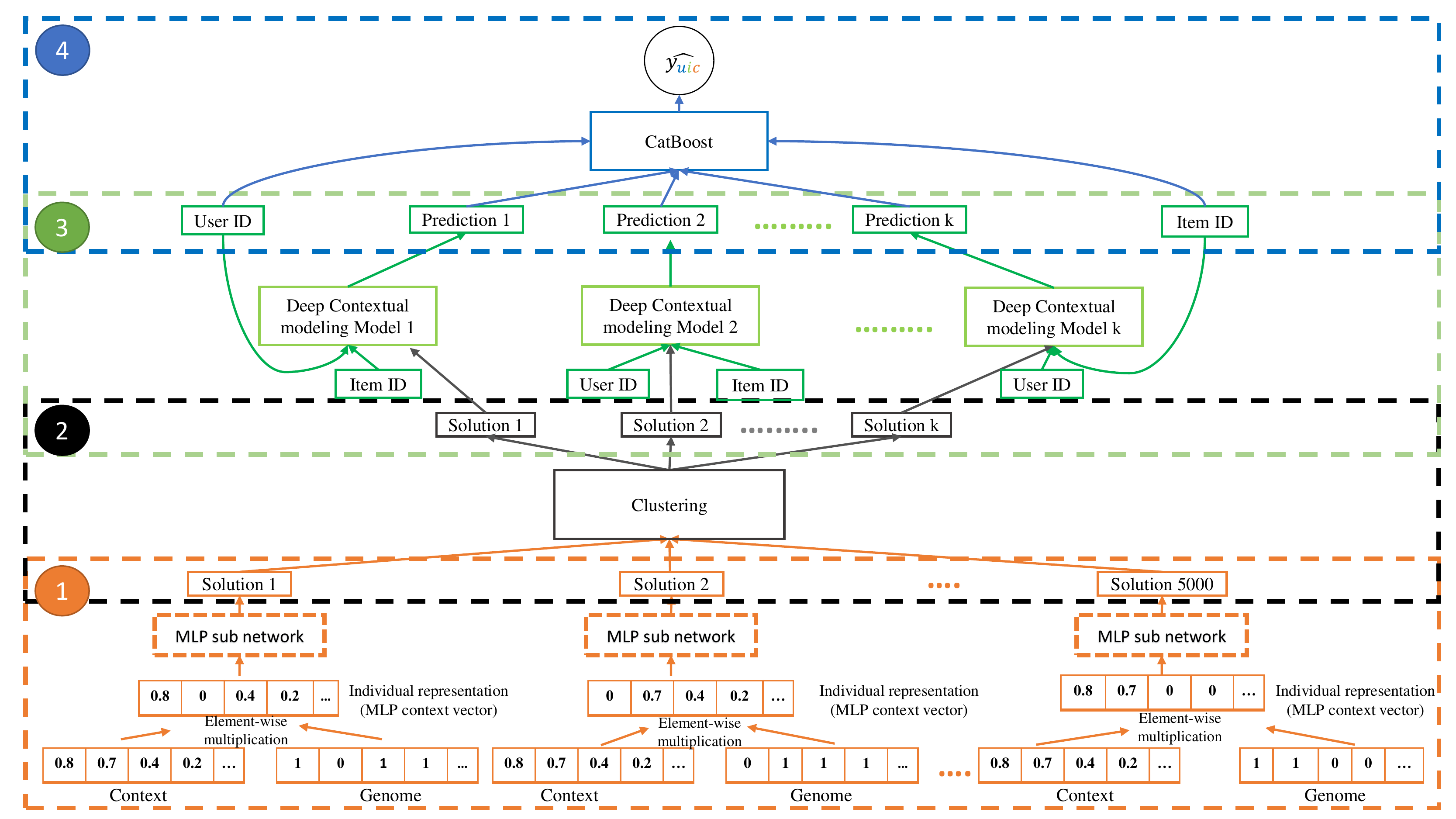}
    \caption{An architecture of the ensemble extension}
    \label{fig:ensemble}
\end{figure}

Thus, we extended our method again, this time by changing the deep contextual model of~\cite{livne2019deep}, as depicted in \autoref{fig:ensemble}. Specifically, we propose the following changes to the architecture, after the evolutionary process terminates:
\begin{enumerate}
\item Select top \emph{unique} 5K individuals according to their AUC performance.
\item Cluster these individuals (we used the DBSCAN clustering algorithm~\cite{ester1996density}). Then select the top individual, in terms of AUC, from each cluster. To compute the similarity between each pair of individuals, we used the Jaccard similarity index.
\item Train the complete deep contextual model of~\cite{livne2019deep} (including user and item) for each of these individuals.
\item Finally, ensemble the resulted models (we used stacking~\cite{hastie2009elements} with CatBoost~\cite{prokhorenkova2018catboost} meta-learner, a state-of-the-art boosting algorithm).
\end{enumerate}

\subsection{Extension \#2 --- Results}
In tables \ref{tab:catboost_solution_cars} and \ref{tab:catboost_solution_Hearo} we compare the performance of the ensemble extension to the performance of the previous solutions and baselines.

Our hypothesis was correct, and this extension outperformed all other solutions and baselines with a margin in the AUC and the log-loss metrics. This extension takes advantage of the already generated individuals and utilizes them for improving the accuracy while preserving the low number of contextual dimensions. 

To validate the significance of our results compared to each of the baselines, we performed statistical tests. We first used the adjusted Friedman test in order to reject the null hypothesis that all classifiers performed the same. The statistical test rejected the null hypothesis with a confidence level of 95\% for both datasets. We then used the Bonferroni–Dunn test to examine whether our solution algorithm performed significantly better than the existing baselines. According to the test, our ensemble solutions statistically outperformed all baselines with a 99\% confidence level for both datasets, except for the FSCM and the LNCM baselines in Hearo, where the significant confidence level was 95\%. 

\begin{table}[ht]
\centering
\begin{tabular}{ll|>{\centering}p{1.7cm}|>{\centering}p{1.7cm}|>{\centering}p{1.1cm}|c|c}
Type & Algorithm & \# Contextual Dimensions & \# Contextual Features & Explicit? (V/X) & AUC & Log loss \tabularnewline \hline
% Best Explicit Baseline & NeuMF & 0 & 0 & V& 0.7953 & 0.481 \\ \hline
% Best Latent Baseline & ENCM & 2 & 6 & V& 0.7925 & 0.482 \\
% With Feature Selection & FSCM & 5 & 9 & V&0.7989 & 0.459 \\
Best explicit baseline & BSCM & 14 & 478 & V& 0.8056 & 0.448  \\ \hline
% Deep Context Aware Models & LNCM & 14 &480 &X & 0.7993 & 0.454  \\
Best latent baseline & SLCM& 14 &480 & X& 0.8004 & 0.454  \\ \bottomrule
% Deep Context Aware Models & NCM & 14 &480 & V& 0.8024 & 0.451  \\
% Without Feature Reduction & DeepFM & 14 &480 &V& 0.7595 & 0.499  \\
% or Selection & FiBiNET & 14 &480&V & 0.7623 & 0.517 \\ \bottomrule
Basic solution & Full trained & 12 & 88 &V & 0.8094 & 0.443  \\
 & Predict only & 11 & 51 & V&0.7997 & 0.460  \\ \hline
Extension \#1 --- & Fully trained & 9 & 81 & V& 0.8075 & 0.445  \\
minimizing dimensions & Predict only & 5 & 34 & V& 0.7984 & 0.460  \\ \hline
Extension \#2 --- & Full trained & 13 & 127 & V&0.8404 & 0.414 \\
Ensemble w/ basic sol. & Predict only & 11 & 72 & V&0.8387 & 0.415  \\ \hline
Extension \#2 --- & Fully trained & \textbf{9} & \textbf{109} & V&\textbf{0.8407} & \textbf{0.413}  \\
Ensemble w/ ext. \#1 & Predict only & \textbf{5} & \textbf{55} &V& \textbf{0.8388} & \textbf{0.415}  \\
\hline
\end{tabular}
\vspace{1ex}
\caption{A comparison of all methods and baselines --- CARS dataset.}
\label{tab:catboost_solution_cars}
\end{table}

\begin{table}[ht]
\centering
\begin{tabular}{ll|>{\centering}p{1.7cm}|>{\centering}p{1.7cm}|>{\centering}p{1.1cm}|c|c}
Type & Algorithm & \# Contextual Dimensions & \# Contextual Features & Explicit? (V/X) & AUC & Log loss \tabularnewline \hline
% No Contextual Features Models & NeuMF & 0 & 0 & V& 0.8007 & 0.542  \\ \hline
% Deep Context Aware Models & ENCM & 2 & 17 & V&0.8177 & 0.556  \\
% With Feature Selection & FSCM & 5 & 6 &V& 0.8164 & 0.529  \\
%     & BSCM & 16 & 658 &V& 0.8160 & 0.526  \\ \hline
% Deep Context Aware Models & LNCM & 16 & 661 & X& 0.8095 & 0.532  \\

Best explicit baseline & NCM & 16 & 661 & V& 0.8191 & 0.519  \\ \hline
Best latent baseline & SLCM & 16 & 661 & X& 0.8203 & 0.518 \\ \bottomrule
% Without Feature Reduction & DeepFM & 16 & 661 &V & 0.7178 & 0.652  \\
% or Selection & FiBiNET & 16 & 661 &V & 0.7294 & 0.646 \\ \bottomrule
Basic solution & Fully trained & 16 & 96 & V& 0.8162 & 0.520  \\
 & Predict only & 15 & 136 & V& 0.8228 & 0.516  \\ \hline
Extension \#1 --- & Fully trained & 9 & 94 & V & 0.8108 & 0.527  \\
minimizing dimensions & Predict only & 8 & 123 & V & 0.8154 & 0.521  \\ \hline
Extension \#2 --- & Fully trained & 16 & 96 & V & 0.8395 & 0.501  \\
Ensemble w/ basic sol. & Predict only & 15 & 142 & V&\textbf{0.8414} & \textbf{0.494}  \\ \hline
Extension \#2 --- & Fully trained & \textbf{9} & \textbf{118} & V & \textbf{0.8395} & \textbf{0.495}  \\
Ensemble w/ ext. \#1 & Predict only & \textbf{8} & \textbf{148} & V & \textbf{0.8370} & \textbf{0.496}  \\
\hline
\end{tabular}
\vspace{1ex}
\caption{A comparison of all methods and baselines --- Hearo dataset.}
\label{tab:catboost_solution_Hearo}
\end{table}

\section{Conclusions and Future Work} \label{section:Conclusions}
In this paper, we propose a novel approach for selecting low-dimensional subsets of contextual information and incorporating them within CARS. Our method outperforms state-of-the-art CARS algorithms and addresses user aspects, such as transparency, battery consumption and privacy. 

In future work, we believe that the approach can be generalized to additional recommendation tasks such as ranking and regression (i.e., top-N recommendation and rating prediction). In addition, we believe that the first extensions can be further improved by applying weights for each contextual dimension. For example, GPS sensor can be assigned with a higher weight than the Wi-Fi sensor, when considering battery consumption and privacy issues.

\begin{acks}
This research was partially supported by the Israeli Council for Higher Education (CHE) via Data Science Research Center, Ben-Gurion University of the Negev, Israel.
\end{acks}

\bibliographystyle{main}
\bibliography{main}

%%% -*-BibTeX-*-
%%% Do NOT edit. File created by BibTeX with style
%%% ACM-Reference-Format-Journals [18-Jan-2012].

\begin{thebibliography}{44}

%%% ====================================================================
%%% NOTE TO THE USER: you can override these defaults by providing
%%% customized versions of any of these macros before the \bibliography
%%% command.  Each of them MUST provide its own final punctuation,
%%% except for \shownote{}, \showDOI{}, and \showURL{}.  The latter two
%%% do not use final punctuation, in order to avoid confusing it with
%%% the Web address.
%%%
%%% To suppress output of a particular field, define its macro to expand
%%% to an empty string, or better, \unskip, like this:
%%%
%%% \newcommand{\showDOI}[1]{\unskip}   % LaTeX syntax
%%%
%%% \def \showDOI #1{\unskip}           % plain TeX syntax
%%%
%%% ====================================================================

\ifx \showCODEN    \undefined \def \showCODEN     #1{\unskip}     \fi
\ifx \showDOI      \undefined \def \showDOI       #1{#1}\fi
\ifx \showISBNx    \undefined \def \showISBNx     #1{\unskip}     \fi
\ifx \showISBNxiii \undefined \def \showISBNxiii  #1{\unskip}     \fi
\ifx \showISSN     \undefined \def \showISSN      #1{\unskip}     \fi
\ifx \showLCCN     \undefined \def \showLCCN      #1{\unskip}     \fi
\ifx \shownote     \undefined \def \shownote      #1{#1}          \fi
\ifx \showarticletitle \undefined \def \showarticletitle #1{#1}   \fi
\ifx \showURL      \undefined \def \showURL       {\relax}        \fi
% The following commands are used for tagged output and should be
% invisible to TeX
\providecommand\bibfield[2]{#2}
\providecommand\bibinfo[2]{#2}
\providecommand\natexlab[1]{#1}
\providecommand\showeprint[2][]{arXiv:#2}

\bibitem[\protect\citeauthoryear{Adomavicius, Sankaranarayanan, Sen, and
  Tuzhilin}{Adomavicius et~al\mbox{.}}{2005}]%
        {adomavicius2005incorporating}
\bibfield{author}{\bibinfo{person}{Gediminas Adomavicius},
  \bibinfo{person}{Ramesh Sankaranarayanan}, \bibinfo{person}{Shahana Sen},
  {and} \bibinfo{person}{Alexander Tuzhilin}.} \bibinfo{year}{2005}\natexlab{}.
\newblock \showarticletitle{Incorporating contextual information in recommender
  systems using a multidimensional approach}.
\newblock \bibinfo{journal}{\emph{ACM Transactions on Information Systems
  (TOIS)}} \bibinfo{volume}{23}, \bibinfo{number}{1} (\bibinfo{year}{2005}),
  \bibinfo{pages}{103--145}.
\newblock


\bibitem[\protect\citeauthoryear{Adomavicius and Tuzhilin}{Adomavicius and
  Tuzhilin}{2011}]%
        {adomavicius2011context}
\bibfield{author}{\bibinfo{person}{Gediminas Adomavicius} {and}
  \bibinfo{person}{Alexander Tuzhilin}.} \bibinfo{year}{2011}\natexlab{}.
\newblock \showarticletitle{Context-Aware Recommender Systems}.
\newblock In \bibinfo{booktitle}{\emph{Recommender Systems Handbook}},
  \bibfield{editor}{\bibinfo{person}{Francesco Ricci}, \bibinfo{person}{Lior
  Rokach}, \bibinfo{person}{Bracha Shapira}, {and} \bibinfo{person}{Paul~B.
  Kantor}} (Eds.). \bibinfo{publisher}{Springer US}, \bibinfo{address}{Boston,
  MA}, \bibinfo{pages}{217--253}.
\newblock
\showISBNx{978-0-387-85820-3}
\urldef\tempurl%
\url{https://doi.org/10.1007/978-0-387-85820-3_7}
\showDOI{\tempurl}


\bibitem[\protect\citeauthoryear{Aggarwal}{Aggarwal}{2016}]%
        {aggarwal2016recommender}
\bibfield{author}{\bibinfo{person}{Charu~C Aggarwal}.}
  \bibinfo{year}{2016}\natexlab{}.
\newblock \bibinfo{booktitle}{\emph{Recommender systems}}.
\newblock \bibinfo{publisher}{Springer US}, \bibinfo{address}{Boston, MA}.
\newblock


\bibitem[\protect\citeauthoryear{Baltrunas, Ludwig, and Ricci}{Baltrunas
  et~al\mbox{.}}{2011}]%
        {baltrunas2011matrix}
\bibfield{author}{\bibinfo{person}{Linas Baltrunas}, \bibinfo{person}{Bernd
  Ludwig}, {and} \bibinfo{person}{Francesco Ricci}.}
  \bibinfo{year}{2011}\natexlab{}.
\newblock \showarticletitle{Matrix Factorization Techniques for Context Aware
  Recommendation}. In \bibinfo{booktitle}{\emph{Proceedings of the Fifth ACM
  Conference on Recommender Systems}} (Chicago, Illinois, USA)
  \emph{(\bibinfo{series}{RecSys ’11})}. \bibinfo{publisher}{Association for
  Computing Machinery}, \bibinfo{address}{New York, NY, USA},
  \bibinfo{pages}{301–304}.
\newblock
\showISBNx{9781450306836}
\urldef\tempurl%
\url{https://doi.org/10.1145/2043932.2043988}
\showDOI{\tempurl}


\bibitem[\protect\citeauthoryear{Ben~Abdesslem, Phillips, and
  Henderson}{Ben~Abdesslem et~al\mbox{.}}{2009}]%
        {ben2009less}
\bibfield{author}{\bibinfo{person}{Fehmi Ben~Abdesslem},
  \bibinfo{person}{Andrew Phillips}, {and} \bibinfo{person}{Tristan
  Henderson}.} \bibinfo{year}{2009}\natexlab{}.
\newblock \showarticletitle{Less is More: Energy-Efficient Mobile Sensing with
  Senseless}. In \bibinfo{booktitle}{\emph{Proceedings of the 1st ACM Workshop
  on Networking, Systems, and Applications for Mobile Handhelds}} (Barcelona,
  Spain) \emph{(\bibinfo{series}{MobiHeld ’09})}.
  \bibinfo{publisher}{Association for Computing Machinery},
  \bibinfo{address}{New York, NY, USA}, \bibinfo{pages}{61–62}.
\newblock
\showISBNx{9781605584447}
\urldef\tempurl%
\url{https://doi.org/10.1145/1592606.1592621}
\showDOI{\tempurl}


\bibitem[\protect\citeauthoryear{Chandrashekar and Sahin}{Chandrashekar and
  Sahin}{2014}]%
        {chandrashekar2014survey}
\bibfield{author}{\bibinfo{person}{Girish Chandrashekar} {and}
  \bibinfo{person}{Ferat Sahin}.} \bibinfo{year}{2014}\natexlab{}.
\newblock \showarticletitle{A survey on feature selection methods}.
\newblock \bibinfo{journal}{\emph{Computers \& Electrical Engineering}}
  \bibinfo{volume}{40}, \bibinfo{number}{1} (\bibinfo{year}{2014}),
  \bibinfo{pages}{16--28}.
\newblock


\bibitem[\protect\citeauthoryear{Chen and Hsieh}{Chen and Hsieh}{2012}]%
        {chen2012personalized}
\bibfield{author}{\bibinfo{person}{Peng-Ting Chen} {and}
  \bibinfo{person}{Hsin-Pei Hsieh}.} \bibinfo{year}{2012}\natexlab{}.
\newblock \showarticletitle{Personalized mobile advertising: Its key
  attributes, trends, and social impact}.
\newblock \bibinfo{journal}{\emph{Technological Forecasting and Social Change}}
  \bibinfo{volume}{79}, \bibinfo{number}{3} (\bibinfo{year}{2012}),
  \bibinfo{pages}{543--557}.
\newblock


\bibitem[\protect\citeauthoryear{Chollet et~al\mbox{.}}{Chollet
  et~al\mbox{.}}{2015}]%
        {chollet2015keras}
\bibfield{author}{\bibinfo{person}{Fran\c{c}ois Chollet} {et~al\mbox{.}}}
  \bibinfo{year}{2015}\natexlab{}.
\newblock \bibinfo{title}{Keras}.
\newblock \bibinfo{howpublished}{\url{https://keras.io}}.
\newblock


\bibitem[\protect\citeauthoryear{Consolvo, McDonald, Toscos, Chen, Froehlich,
  Harrison, Klasnja, LaMarca, LeGrand, Libby, Smith, and Landay}{Consolvo
  et~al\mbox{.}}{2008}]%
        {consolvo2008activity}
\bibfield{author}{\bibinfo{person}{Sunny Consolvo}, \bibinfo{person}{David~W.
  McDonald}, \bibinfo{person}{Tammy Toscos}, \bibinfo{person}{Mike~Y. Chen},
  \bibinfo{person}{Jon Froehlich}, \bibinfo{person}{Beverly Harrison},
  \bibinfo{person}{Predrag Klasnja}, \bibinfo{person}{Anthony LaMarca},
  \bibinfo{person}{Louis LeGrand}, \bibinfo{person}{Ryan Libby},
  \bibinfo{person}{Ian Smith}, {and} \bibinfo{person}{James~A. Landay}.}
  \bibinfo{year}{2008}\natexlab{}.
\newblock \showarticletitle{Activity Sensing in the Wild: A Field Trial of
  Ubifit Garden}. In \bibinfo{booktitle}{\emph{Proceedings of the SIGCHI
  Conference on Human Factors in Computing Systems}} (Florence, Italy)
  \emph{(\bibinfo{series}{CHI ’08})}. \bibinfo{publisher}{Association for
  Computing Machinery}, \bibinfo{address}{New York, NY, USA},
  \bibinfo{pages}{1797–1806}.
\newblock
\showISBNx{9781605580111}
\urldef\tempurl%
\url{https://doi.org/10.1145/1357054.1357335}
\showDOI{\tempurl}


\bibitem[\protect\citeauthoryear{Costa and Dolog}{Costa and Dolog}{2019}]%
        {costa2019collective}
\bibfield{author}{\bibinfo{person}{Felipe Soares~da Costa} {and}
  \bibinfo{person}{Peter Dolog}.} \bibinfo{year}{2019}\natexlab{}.
\newblock \showarticletitle{Collective Embedding for Neural Context-Aware
  Recommender Systems}. In \bibinfo{booktitle}{\emph{Proceedings of the 13th
  ACM Conference on Recommender Systems}} (Copenhagen, Denmark)
  \emph{(\bibinfo{series}{RecSys ’19})}. \bibinfo{publisher}{Association for
  Computing Machinery}, \bibinfo{address}{New York, NY, USA},
  \bibinfo{pages}{201–209}.
\newblock
\showISBNx{9781450362436}
\urldef\tempurl%
\url{https://doi.org/10.1145/3298689.3347028}
\showDOI{\tempurl}


\bibitem[\protect\citeauthoryear{Cramer, Evers, Ramlal, Van~Someren, Rutledge,
  Stash, Aroyo, and Wielinga}{Cramer et~al\mbox{.}}{2008}]%
        {cramer2008effects}
\bibfield{author}{\bibinfo{person}{Henriette Cramer}, \bibinfo{person}{Vanessa
  Evers}, \bibinfo{person}{Satyan Ramlal}, \bibinfo{person}{Maarten
  Van~Someren}, \bibinfo{person}{Lloyd Rutledge}, \bibinfo{person}{Natalia
  Stash}, \bibinfo{person}{Lora Aroyo}, {and} \bibinfo{person}{Bob Wielinga}.}
  \bibinfo{year}{2008}\natexlab{}.
\newblock \showarticletitle{The effects of transparency on trust in and
  acceptance of a content-based art recommender}.
\newblock \bibinfo{journal}{\emph{User Modeling and User-Adapted Interaction}}
  \bibinfo{volume}{18}, \bibinfo{number}{5} (\bibinfo{year}{2008}),
  \bibinfo{pages}{455}.
\newblock


\bibitem[\protect\citeauthoryear{Eiben, Smith, et~al\mbox{.}}{Eiben
  et~al\mbox{.}}{2003}]%
        {eiben2003introduction}
\bibfield{author}{\bibinfo{person}{Agoston~E Eiben}, \bibinfo{person}{James~E
  Smith}, {et~al\mbox{.}}} \bibinfo{year}{2003}\natexlab{}.
\newblock \bibinfo{booktitle}{\emph{Introduction to evolutionary computing}}.
  Vol.~\bibinfo{volume}{53}.
\newblock \bibinfo{publisher}{Springer}.
\newblock


\bibitem[\protect\citeauthoryear{Ester, Kriegel, Sander, and Xu}{Ester
  et~al\mbox{.}}{1996}]%
        {ester1996density}
\bibfield{author}{\bibinfo{person}{Martin Ester}, \bibinfo{person}{Hans-Peter
  Kriegel}, \bibinfo{person}{J\"{o}rg Sander}, {and} \bibinfo{person}{Xiaowei
  Xu}.} \bibinfo{year}{1996}\natexlab{}.
\newblock \showarticletitle{A Density-Based Algorithm for Discovering Clusters
  in Large Spatial Databases with Noise}. In
  \bibinfo{booktitle}{\emph{Proceedings of the Second International Conference
  on Knowledge Discovery and Data Mining}} (Portland, Oregon)
  \emph{(\bibinfo{series}{KDD’96})}. \bibinfo{publisher}{AAAI Press},
  \bibinfo{pages}{226–231}.
\newblock


\bibitem[\protect\citeauthoryear{Fortin, {De Rainville}, Gardner, Parizeau, and
  Gagn\'e}{Fortin et~al\mbox{.}}{2012}]%
        {DEAP_JMLR2012}
\bibfield{author}{\bibinfo{person}{F\'elix-Antoine Fortin},
  \bibinfo{person}{Fran\c{c}ois-Michel {De Rainville}},
  \bibinfo{person}{Marc-Andr\'e Gardner}, \bibinfo{person}{Marc Parizeau},
  {and} \bibinfo{person}{Christian Gagn\'e}.} \bibinfo{year}{2012}\natexlab{}.
\newblock \showarticletitle{{DEAP}: Evolutionary Algorithms Made Easy}.
\newblock \bibinfo{journal}{\emph{Journal of Machine Learning Research}}
  \bibinfo{volume}{13} (\bibinfo{date}{jul} \bibinfo{year}{2012}),
  \bibinfo{pages}{2171--2175}.
\newblock


\bibitem[\protect\citeauthoryear{Guo, Tang, Ye, Li, and He}{Guo
  et~al\mbox{.}}{2017}]%
        {guo2017deepfm}
\bibfield{author}{\bibinfo{person}{Huifeng Guo}, \bibinfo{person}{Ruiming
  Tang}, \bibinfo{person}{Yunming Ye}, \bibinfo{person}{Zhenguo Li}, {and}
  \bibinfo{person}{Xiuqiang He}.} \bibinfo{year}{2017}\natexlab{}.
\newblock \showarticletitle{DeepFM: a factorization-machine based neural
  network for CTR prediction}.
\newblock \bibinfo{journal}{\emph{arXiv preprint arXiv:1703.04247}}
  \bibinfo{volume}{24}, \bibinfo{number}{3} (\bibinfo{year}{2017}),
  \bibinfo{pages}{262--290}.
\newblock


\bibitem[\protect\citeauthoryear{Hastie, Tibshirani, and Friedman}{Hastie
  et~al\mbox{.}}{2009}]%
        {hastie2009elements}
\bibfield{author}{\bibinfo{person}{Trevor Hastie}, \bibinfo{person}{Robert
  Tibshirani}, {and} \bibinfo{person}{Jerome Friedman}.}
  \bibinfo{year}{2009}\natexlab{}.
\newblock \bibinfo{booktitle}{\emph{The elements of statistical learning: data
  mining, inference, and prediction}}.
\newblock \bibinfo{publisher}{Springer Science \& Business Media}.
\newblock


\bibitem[\protect\citeauthoryear{He, Liao, Zhang, Nie, Hu, and Chua}{He
  et~al\mbox{.}}{2017}]%
        {he2017neural}
\bibfield{author}{\bibinfo{person}{Xiangnan He}, \bibinfo{person}{Lizi Liao},
  \bibinfo{person}{Hanwang Zhang}, \bibinfo{person}{Liqiang Nie},
  \bibinfo{person}{Xia Hu}, {and} \bibinfo{person}{Tat-Seng Chua}.}
  \bibinfo{year}{2017}\natexlab{}.
\newblock \showarticletitle{Neural Collaborative Filtering}. In
  \bibinfo{booktitle}{\emph{Proceedings of the 26th International Conference on
  World Wide Web}} (Perth, Australia) \emph{(\bibinfo{series}{WWW ’17})}.
  \bibinfo{publisher}{International World Wide Web Conferences Steering
  Committee}, \bibinfo{address}{Republic and Canton of Geneva, CHE},
  \bibinfo{pages}{173–182}.
\newblock
\showISBNx{9781450349130}
\urldef\tempurl%
\url{https://doi.org/10.1145/3038912.3052569}
\showDOI{\tempurl}


\bibitem[\protect\citeauthoryear{Huang, Zhang, and Zhang}{Huang
  et~al\mbox{.}}{2019}]%
        {huang2019fibinet}
\bibfield{author}{\bibinfo{person}{Tongwen Huang}, \bibinfo{person}{Zhiqi
  Zhang}, {and} \bibinfo{person}{Junlin Zhang}.}
  \bibinfo{year}{2019}\natexlab{}.
\newblock \showarticletitle{FiBiNET: Combining Feature Importance and Bilinear
  Feature Interaction for Click-through Rate Prediction}. In
  \bibinfo{booktitle}{\emph{Proceedings of the 13th ACM Conference on
  Recommender Systems}} (Copenhagen, Denmark) \emph{(\bibinfo{series}{RecSys
  ’19})}. \bibinfo{publisher}{Association for Computing Machinery},
  \bibinfo{address}{New York, NY, USA}, \bibinfo{pages}{169–177}.
\newblock
\showISBNx{9781450362436}
\urldef\tempurl%
\url{https://doi.org/10.1145/3298689.3347043}
\showDOI{\tempurl}


\bibitem[\protect\citeauthoryear{Jiang, Chin, Wang, Qu, and Tsui}{Jiang
  et~al\mbox{.}}{2017}]%
        {jiang2017modified}
\bibfield{author}{\bibinfo{person}{Shancheng Jiang}, \bibinfo{person}{Kwai-Sang
  Chin}, \bibinfo{person}{Long Wang}, \bibinfo{person}{Gang Qu}, {and}
  \bibinfo{person}{Kwok~L Tsui}.} \bibinfo{year}{2017}\natexlab{}.
\newblock \showarticletitle{Modified genetic algorithm-based feature selection
  combined with pre-trained deep neural network for demand forecasting in
  outpatient department}.
\newblock \bibinfo{journal}{\emph{Expert systems with applications}}
  \bibinfo{volume}{82} (\bibinfo{year}{2017}), \bibinfo{pages}{216--230}.
\newblock


\bibitem[\protect\citeauthoryear{Karatzoglou, Amatriain, Baltrunas, and
  Oliver}{Karatzoglou et~al\mbox{.}}{2010}]%
        {karatzoglou2010multiverse}
\bibfield{author}{\bibinfo{person}{Alexandros Karatzoglou},
  \bibinfo{person}{Xavier Amatriain}, \bibinfo{person}{Linas Baltrunas}, {and}
  \bibinfo{person}{Nuria Oliver}.} \bibinfo{year}{2010}\natexlab{}.
\newblock \showarticletitle{Multiverse Recommendation: N-Dimensional Tensor
  Factorization for Context-Aware Collaborative Filtering}. In
  \bibinfo{booktitle}{\emph{Proceedings of the Fourth ACM Conference on
  Recommender Systems}} (Barcelona, Spain) \emph{(\bibinfo{series}{RecSys
  ’10})}. \bibinfo{publisher}{Association for Computing Machinery},
  \bibinfo{address}{New York, NY, USA}, \bibinfo{pages}{79–86}.
\newblock
\showISBNx{9781605589060}
\urldef\tempurl%
\url{https://doi.org/10.1145/1864708.1864727}
\showDOI{\tempurl}


\bibitem[\protect\citeauthoryear{Kim, Park, Oh, Lee, and Yu}{Kim
  et~al\mbox{.}}{2016}]%
        {kim2016convolutional}
\bibfield{author}{\bibinfo{person}{Donghyun Kim}, \bibinfo{person}{Chanyoung
  Park}, \bibinfo{person}{Jinoh Oh}, \bibinfo{person}{Sungyoung Lee}, {and}
  \bibinfo{person}{Hwanjo Yu}.} \bibinfo{year}{2016}\natexlab{}.
\newblock \showarticletitle{Convolutional Matrix Factorization for Document
  Context-Aware Recommendation}. In \bibinfo{booktitle}{\emph{Proceedings of
  the 10th ACM Conference on Recommender Systems}} (Boston, Massachusetts, USA)
  \emph{(\bibinfo{series}{RecSys ’16})}. \bibinfo{publisher}{Association for
  Computing Machinery}, \bibinfo{address}{New York, NY, USA},
  \bibinfo{pages}{233–240}.
\newblock
\showISBNx{9781450340359}
\urldef\tempurl%
\url{https://doi.org/10.1145/2959100.2959165}
\showDOI{\tempurl}


\bibitem[\protect\citeauthoryear{Kingma and Ba}{Kingma and Ba}{2014}]%
        {kingma2014adam}
\bibfield{author}{\bibinfo{person}{Diederik~P Kingma} {and}
  \bibinfo{person}{Jimmy Ba}.} \bibinfo{year}{2014}\natexlab{}.
\newblock \showarticletitle{Adam: A method for stochastic optimization}.
\newblock \bibinfo{journal}{\emph{arXiv preprint arXiv:1412.6980}}
  (\bibinfo{year}{2014}).
\newblock


\bibitem[\protect\citeauthoryear{Kwapisz, Weiss, and Moore}{Kwapisz
  et~al\mbox{.}}{2011}]%
        {kwapisz2011activity}
\bibfield{author}{\bibinfo{person}{Jennifer~R Kwapisz}, \bibinfo{person}{Gary~M
  Weiss}, {and} \bibinfo{person}{Samuel~A Moore}.}
  \bibinfo{year}{2011}\natexlab{}.
\newblock \showarticletitle{Activity recognition using cell phone
  accelerometers}.
\newblock \bibinfo{journal}{\emph{ACM SigKDD Explorations Newsletter}}
  \bibinfo{volume}{12}, \bibinfo{number}{2} (\bibinfo{year}{2011}),
  \bibinfo{pages}{74--82}.
\newblock


\bibitem[\protect\citeauthoryear{Liu, Tang, Chen, Yu, Guo, and Zhang}{Liu
  et~al\mbox{.}}{2019}]%
        {liu2019feature}
\bibfield{author}{\bibinfo{person}{Bin Liu}, \bibinfo{person}{Ruiming Tang},
  \bibinfo{person}{Yingzhi Chen}, \bibinfo{person}{Jinkai Yu},
  \bibinfo{person}{Huifeng Guo}, {and} \bibinfo{person}{Yuzhou Zhang}.}
  \bibinfo{year}{2019}\natexlab{}.
\newblock \showarticletitle{Feature Generation by Convolutional Neural Network
  for Click-Through Rate Prediction}. In \bibinfo{booktitle}{\emph{The World
  Wide Web Conference}} (San Francisco, CA, USA) \emph{(\bibinfo{series}{WWW
  ’19})}. \bibinfo{publisher}{Association for Computing Machinery},
  \bibinfo{address}{New York, NY, USA}, \bibinfo{pages}{1119–1129}.
\newblock
\showISBNx{9781450366748}
\urldef\tempurl%
\url{https://doi.org/10.1145/3308558.3313497}
\showDOI{\tempurl}


\bibitem[\protect\citeauthoryear{Liu, Motoda, Setiono, and Zhao}{Liu
  et~al\mbox{.}}{2010}]%
        {liu2010feature}
\bibfield{author}{\bibinfo{person}{Huan Liu}, \bibinfo{person}{Hiroshi Motoda},
  \bibinfo{person}{Rudy Setiono}, {and} \bibinfo{person}{Zheng Zhao}.}
  \bibinfo{year}{2010}\natexlab{}.
\newblock \showarticletitle{Feature selection: An ever evolving frontier in
  data mining}. In \bibinfo{booktitle}{\emph{Feature selection in data
  mining}}. \bibinfo{pages}{4--13}.
\newblock


\bibitem[\protect\citeauthoryear{Liu, Yin, Gao, and Tan}{Liu
  et~al\mbox{.}}{2008}]%
        {liu2008wrapper}
\bibfield{author}{\bibinfo{person}{Yue Liu}, \bibinfo{person}{Yafeng Yin},
  \bibinfo{person}{Junjun Gao}, {and} \bibinfo{person}{Chongli Tan}.}
  \bibinfo{year}{2008}\natexlab{}.
\newblock \showarticletitle{Wrapper feature selection optimized SVM model for
  demand forecasting}. In \bibinfo{booktitle}{\emph{2008 The 9th International
  Conference for Young Computer Scientists}}. IEEE, \bibinfo{pages}{953--958}.
\newblock


\bibitem[\protect\citeauthoryear{Livne, Unger, Shapira, and Rokach}{Livne
  et~al\mbox{.}}{2019}]%
        {livne2019deep}
\bibfield{author}{\bibinfo{person}{Amit Livne}, \bibinfo{person}{Moshe Unger},
  \bibinfo{person}{Bracha Shapira}, {and} \bibinfo{person}{Lior Rokach}.}
  \bibinfo{year}{2019}\natexlab{}.
\newblock \showarticletitle{Deep Context-Aware Recommender System Utilizing
  Sequential Latent Context}.
\newblock \bibinfo{journal}{\emph{arXiv preprint arXiv:1909.03999}}
  (\bibinfo{year}{2019}).
\newblock


\bibitem[\protect\citeauthoryear{Lu, Pan, Lane, Choudhury, and Campbell}{Lu
  et~al\mbox{.}}{2009}]%
        {lu2009soundsense}
\bibfield{author}{\bibinfo{person}{Hong Lu}, \bibinfo{person}{Wei Pan},
  \bibinfo{person}{Nicholas~D. Lane}, \bibinfo{person}{Tanzeem Choudhury},
  {and} \bibinfo{person}{Andrew~T. Campbell}.} \bibinfo{year}{2009}\natexlab{}.
\newblock \showarticletitle{SoundSense: Scalable Sound Sensing for
  People-Centric Applications on Mobile Phones}. In
  \bibinfo{booktitle}{\emph{Proceedings of the 7th International Conference on
  Mobile Systems, Applications, and Services}} (Krak\'{o}w, Poland)
  \emph{(\bibinfo{series}{MobiSys ’09})}. \bibinfo{publisher}{Association for
  Computing Machinery}, \bibinfo{address}{New York, NY, USA},
  \bibinfo{pages}{165–178}.
\newblock
\showISBNx{9781605585666}
\urldef\tempurl%
\url{https://doi.org/10.1145/1555816.1555834}
\showDOI{\tempurl}


\bibitem[\protect\citeauthoryear{Perera, Zaslavsky, Christen, and
  Georgakopoulos}{Perera et~al\mbox{.}}{2014}]%
        {perera2014context}
\bibfield{author}{\bibinfo{person}{Charith Perera}, \bibinfo{person}{Arkady
  Zaslavsky}, \bibinfo{person}{Peter Christen}, {and}
  \bibinfo{person}{Dimitrios Georgakopoulos}.} \bibinfo{year}{2014}\natexlab{}.
\newblock \showarticletitle{Context aware computing for the internet of things:
  A survey}.
\newblock \bibinfo{journal}{\emph{IEEE communications surveys \& tutorials}}
  \bibinfo{volume}{16}, \bibinfo{number}{1} (\bibinfo{year}{2014}),
  \bibinfo{pages}{414--454}.
\newblock


\bibitem[\protect\citeauthoryear{Prokhorenkova, Gusev, Vorobev, Dorogush, and
  Gulin}{Prokhorenkova et~al\mbox{.}}{2018}]%
        {prokhorenkova2018catboost}
\bibfield{author}{\bibinfo{person}{Liudmila Prokhorenkova},
  \bibinfo{person}{Gleb Gusev}, \bibinfo{person}{Aleksandr Vorobev},
  \bibinfo{person}{Anna~Veronika Dorogush}, {and} \bibinfo{person}{Andrey
  Gulin}.} \bibinfo{year}{2018}\natexlab{}.
\newblock \showarticletitle{CatBoost: Unbiased Boosting with Categorical
  Features}. In \bibinfo{booktitle}{\emph{Proceedings of the 32nd International
  Conference on Neural Information Processing Systems}} (Montr\'{e}al, Canada)
  \emph{(\bibinfo{series}{NIPS’18})}. \bibinfo{publisher}{Curran Associates
  Inc.}, \bibinfo{address}{Red Hook, NY, USA}, \bibinfo{pages}{6639–6649}.
\newblock


\bibitem[\protect\citeauthoryear{Saiph~Savage, Baranski, Elva~Chavez, and
  H{\"o}llerer}{Saiph~Savage et~al\mbox{.}}{2012}]%
        {savage2012m}
\bibfield{author}{\bibinfo{person}{Norma Saiph~Savage}, \bibinfo{person}{Maciej
  Baranski}, \bibinfo{person}{Norma Elva~Chavez}, {and} \bibinfo{person}{Tobias
  H{\"o}llerer}.} \bibinfo{year}{2012}\natexlab{}.
\newblock \showarticletitle{I'm feeling LoCo: A Location Based Context Aware
  Recommendation System}.
\newblock In \bibinfo{booktitle}{\emph{Advances in Location-Based Services: 8th
  International Symposium on Location-Based Services, Vienna 2011}},
  \bibfield{editor}{\bibinfo{person}{Georg Gartner} {and}
  \bibinfo{person}{Felix Ortag}} (Eds.). \bibinfo{publisher}{Springer Berlin
  Heidelberg}, \bibinfo{address}{Berlin, Heidelberg}, \bibinfo{pages}{37--54}.
\newblock
\showISBNx{978-3-642-24198-7}
\urldef\tempurl%
\url{https://doi.org/10.1007/978-3-642-24198-7_3}
\showDOI{\tempurl}


\bibitem[\protect\citeauthoryear{Santos, Cardoso, Ferreira, Diniz, and
  Cha{\'\i}nho}{Santos et~al\mbox{.}}{2010}]%
        {santos2010providing}
\bibfield{author}{\bibinfo{person}{Andr{\'e}~C Santos},
  \bibinfo{person}{Jo{\~a}o~MP Cardoso}, \bibinfo{person}{Diogo~R Ferreira},
  \bibinfo{person}{Pedro~C Diniz}, {and} \bibinfo{person}{Paulo Cha{\'\i}nho}.}
  \bibinfo{year}{2010}\natexlab{}.
\newblock \showarticletitle{Providing user context for mobile and social
  networking applications}.
\newblock \bibinfo{journal}{\emph{Pervasive and Mobile Computing}}
  \bibinfo{volume}{6}, \bibinfo{number}{3} (\bibinfo{year}{2010}),
  \bibinfo{pages}{324--341}.
\newblock


\bibitem[\protect\citeauthoryear{Sinha and Swearingen}{Sinha and
  Swearingen}{2002}]%
        {sinha2002role}
\bibfield{author}{\bibinfo{person}{Rashmi Sinha} {and} \bibinfo{person}{Kirsten
  Swearingen}.} \bibinfo{year}{2002}\natexlab{}.
\newblock \showarticletitle{The Role of Transparency in Recommender Systems}.
  In \bibinfo{booktitle}{\emph{CHI ’02 Extended Abstracts on Human Factors in
  Computing Systems}} (Minneapolis, Minnesota, USA) \emph{(\bibinfo{series}{CHI
  EA ’02})}. \bibinfo{publisher}{Association for Computing Machinery},
  \bibinfo{address}{New York, NY, USA}, \bibinfo{pages}{830–831}.
\newblock
\showISBNx{1581134541}
\urldef\tempurl%
\url{https://doi.org/10.1145/506443.506619}
\showDOI{\tempurl}


\bibitem[\protect\citeauthoryear{Sun, Zhang, Tu, and Huang}{Sun
  et~al\mbox{.}}{2013}]%
        {sun2013we}
\bibfield{author}{\bibinfo{person}{Fei Sun}, \bibinfo{person}{Jun Zhang},
  \bibinfo{person}{Lai Tu}, {and} \bibinfo{person}{Benxiong Huang}.}
  \bibinfo{year}{2013}\natexlab{}.
\newblock \showarticletitle{What We Use to Predict a Mobile-Phone Users’
  Status in Campus?}. In \bibinfo{booktitle}{\emph{Proceedings of the 2013 IEEE
  16th International Conference on Computational Science and Engineering}}
  \emph{(\bibinfo{series}{CSE ’13})}. \bibinfo{publisher}{IEEE Computer
  Society}, \bibinfo{address}{USA}, \bibinfo{pages}{1238–1241}.
\newblock
\showISBNx{9780769550961}
\urldef\tempurl%
\url{https://doi.org/10.1109/CSE.2013.184}
\showDOI{\tempurl}


\bibitem[\protect\citeauthoryear{Unger, Bar, Shapira, and Rokach}{Unger
  et~al\mbox{.}}{2016}]%
        {unger2016towards}
\bibfield{author}{\bibinfo{person}{Moshe Unger}, \bibinfo{person}{Ariel Bar},
  \bibinfo{person}{Bracha Shapira}, {and} \bibinfo{person}{Lior Rokach}.}
  \bibinfo{year}{2016}\natexlab{}.
\newblock \showarticletitle{Towards latent context-aware recommendation
  systems}.
\newblock \bibinfo{journal}{\emph{Knowledge-Based Systems}}
  \bibinfo{volume}{104} (\bibinfo{year}{2016}), \bibinfo{pages}{165--178}.
\newblock


\bibitem[\protect\citeauthoryear{Unger, Shapira, Rokach, and Livne}{Unger
  et~al\mbox{.}}{2018}]%
        {unger2018inferring}
\bibfield{author}{\bibinfo{person}{Moshe Unger}, \bibinfo{person}{Bracha
  Shapira}, \bibinfo{person}{Lior Rokach}, {and} \bibinfo{person}{Amit Livne}.}
  \bibinfo{year}{2018}\natexlab{}.
\newblock \showarticletitle{Inferring contextual preferences using deep
  encoder-decoder learners}.
\newblock \bibinfo{journal}{\emph{New Review of Hypermedia and Multimedia}}
  \bibinfo{volume}{24}, \bibinfo{number}{3} (\bibinfo{year}{2018}),
  \bibinfo{pages}{262--290}.
\newblock


\bibitem[\protect\citeauthoryear{Unger and Tuzhilin}{Unger and
  Tuzhilin}{2019}]%
        {unger2019hierarchical}
\bibfield{author}{\bibinfo{person}{Moshe Unger} {and}
  \bibinfo{person}{Alexander Tuzhilin}.} \bibinfo{year}{2019}\natexlab{}.
\newblock \showarticletitle{Hierarchical Latent Context Representation for
  CARS}.
\newblock  (\bibinfo{year}{2019}).
\newblock


\bibitem[\protect\citeauthoryear{Unger, Tuzhilin, and Livne}{Unger
  et~al\mbox{.}}{2020}]%
        {unger2020context}
\bibfield{author}{\bibinfo{person}{Moshe Unger}, \bibinfo{person}{Alexander
  Tuzhilin}, {and} \bibinfo{person}{Amit Livne}.}
  \bibinfo{year}{2020}\natexlab{}.
\newblock \showarticletitle{Context-Aware Recommendations Based on Deep
  Learning Frameworks}.
\newblock \bibinfo{journal}{\emph{ACM Transactions on Management Information
  Systems (TMIS)}} \bibinfo{volume}{11}, \bibinfo{number}{2}
  (\bibinfo{year}{2020}), \bibinfo{pages}{1--15}.
\newblock


\bibitem[\protect\citeauthoryear{Verbert, Manouselis, Ochoa, Wolpers,
  Drachsler, Bosnic, and Duval}{Verbert et~al\mbox{.}}{2012}]%
        {verbert2012context}
\bibfield{author}{\bibinfo{person}{Katrien Verbert}, \bibinfo{person}{Nikos
  Manouselis}, \bibinfo{person}{Xavier Ochoa}, \bibinfo{person}{Martin
  Wolpers}, \bibinfo{person}{Hendrik Drachsler}, \bibinfo{person}{Ivana
  Bosnic}, {and} \bibinfo{person}{Erik Duval}.}
  \bibinfo{year}{2012}\natexlab{}.
\newblock \showarticletitle{Context-aware recommender systems for learning: a
  survey and future challenges}.
\newblock \bibinfo{journal}{\emph{IEEE Transactions on Learning Technologies}}
  \bibinfo{volume}{5}, \bibinfo{number}{4} (\bibinfo{year}{2012}),
  \bibinfo{pages}{318--335}.
\newblock


\bibitem[\protect\citeauthoryear{Wu, Zhao, Zhang, Meng, Zhang, Zhang, and
  Sun}{Wu et~al\mbox{.}}{2017}]%
        {wu2017improving}
\bibfield{author}{\bibinfo{person}{Wenmin Wu}, \bibinfo{person}{Jianli Zhao},
  \bibinfo{person}{Chunsheng Zhang}, \bibinfo{person}{Fang Meng},
  \bibinfo{person}{Zeli Zhang}, \bibinfo{person}{Yang Zhang}, {and}
  \bibinfo{person}{Qiuxia Sun}.} \bibinfo{year}{2017}\natexlab{}.
\newblock \showarticletitle{Improving performance of tensor-based context-aware
  recommenders using Bias Tensor Factorization with context feature
  auto-encoding}.
\newblock \bibinfo{journal}{\emph{Knowledge-Based Systems}}
  \bibinfo{volume}{128} (\bibinfo{year}{2017}), \bibinfo{pages}{71--77}.
\newblock


\bibitem[\protect\citeauthoryear{Xin, Chen, He, Wang, Ding, and Jose}{Xin
  et~al\mbox{.}}{2019}]%
        {xin2019cfm}
\bibfield{author}{\bibinfo{person}{Xin Xin}, \bibinfo{person}{Bo Chen},
  \bibinfo{person}{Xiangnan He}, \bibinfo{person}{Dong Wang},
  \bibinfo{person}{Yue Ding}, {and} \bibinfo{person}{Joemon Jose}.}
  \bibinfo{year}{2019}\natexlab{}.
\newblock \showarticletitle{CFM: Convolutional Factorization Machines for
  Context-Aware Recommendation}. In \bibinfo{booktitle}{\emph{Proceedings of
  the Twenty-Eighth International Joint Conference on Artificial Intelligence,
  {IJCAI-19}}}. \bibinfo{publisher}{International Joint Conferences on
  Artificial Intelligence Organization}, \bibinfo{pages}{3926--3932}.
\newblock
\urldef\tempurl%
\url{https://doi.org/10.24963/ijcai.2019/545}
\showDOI{\tempurl}


\bibitem[\protect\citeauthoryear{Xue, Zhang, Browne, and Yao}{Xue
  et~al\mbox{.}}{2015}]%
        {xue2015survey}
\bibfield{author}{\bibinfo{person}{Bing Xue}, \bibinfo{person}{Mengjie Zhang},
  \bibinfo{person}{Will~N Browne}, {and} \bibinfo{person}{Xin Yao}.}
  \bibinfo{year}{2015}\natexlab{}.
\newblock \showarticletitle{A survey on evolutionary computation approaches to
  feature selection}.
\newblock \bibinfo{journal}{\emph{IEEE Transactions on Evolutionary
  Computation}} \bibinfo{volume}{20}, \bibinfo{number}{4}
  (\bibinfo{year}{2015}), \bibinfo{pages}{606--626}.
\newblock


\bibitem[\protect\citeauthoryear{Zheng, Mobasher, and Burke}{Zheng
  et~al\mbox{.}}{2014}]%
        {zheng2014cslim}
\bibfield{author}{\bibinfo{person}{Yong Zheng}, \bibinfo{person}{Bamshad
  Mobasher}, {and} \bibinfo{person}{Robin Burke}.}
  \bibinfo{year}{2014}\natexlab{}.
\newblock \showarticletitle{CSLIM: Contextual SLIM Recommendation Algorithms}.
  In \bibinfo{booktitle}{\emph{Proceedings of the 8th ACM Conference on
  Recommender Systems}} (Foster City, Silicon Valley, California, USA)
  \emph{(\bibinfo{series}{RecSys ’14})}. \bibinfo{publisher}{Association for
  Computing Machinery}, \bibinfo{address}{New York, NY, USA},
  \bibinfo{pages}{301–304}.
\newblock
\showISBNx{9781450326681}
\urldef\tempurl%
\url{https://doi.org/10.1145/2645710.2645756}
\showDOI{\tempurl}


\bibitem[\protect\citeauthoryear{Zhou, Gu, Zhang, and Fei}{Zhou
  et~al\mbox{.}}{2015}]%
        {zhou2015priori}
\bibfield{author}{\bibinfo{person}{Peng Zhou}, \bibinfo{person}{Xiaojing Gu},
  \bibinfo{person}{Jie Zhang}, {and} \bibinfo{person}{Minrui Fei}.}
  \bibinfo{year}{2015}\natexlab{}.
\newblock \showarticletitle{A priori trust inference with context-aware
  stereotypical deep learning}.
\newblock \bibinfo{journal}{\emph{Knowledge-Based Systems}}
  \bibinfo{volume}{88} (\bibinfo{year}{2015}), \bibinfo{pages}{97--106}.
\newblock


\end{thebibliography}
\end{document}